\documentclass[11pt,twocolumn]{article}

\PassOptionsToPackage{table}{xcolor}

\usepackage[margin=0.75in,top=0.9in,bottom=0.9in]{geometry}
\usepackage{setspace}
\usepackage{amsmath,amssymb,amsfonts}
\usepackage{graphicx}
\usepackage{booktabs}
\usepackage{multirow}
\usepackage{array}
\usepackage{tabularx}
\usepackage{tikz}
\usetikzlibrary{shapes, arrows, positioning, fit, backgrounds}
\usepackage{longtable}
\usepackage{microtype}
\usepackage{enumitem}
\usepackage{algorithm}
\usepackage{algpseudocode}
\usepackage[hyphens]{url}
\usepackage[T1]{fontenc}
\usepackage{xcolor}
\usepackage{float}
\usepackage{subcaption}

\usepackage{mathpazo}           
\usepackage{helvet}             
\usepackage[colorlinks=true,
            linkcolor=blue!50!black,
            citecolor=blue!50!black,
            urlcolor=blue!50!black]{hyperref}
\usepackage[nameinlink,capitalise]{cleveref}

\usepackage{titlesec}
\titleformat{\section}{\large\sffamily\bfseries}{\thesection}{1em}{}
\titleformat{\subsection}{\normalsize\sffamily\bfseries}{\thesubsection}{1em}{}
\titleformat{\subsubsection}{\normalsize\sffamily\itshape}{\thesubsubsection}{1em}{}
\titlespacing*{\section}{0pt}{12pt plus 2pt minus 2pt}{4pt plus 1pt minus 1pt}
\titlespacing*{\subsection}{0pt}{8pt plus 2pt minus 2pt}{2pt plus 1pt minus 1pt}

\usepackage[font={small,sf},labelfont={sf,bf}]{caption}

\usepackage[switch]{lineno}

\begin{document}

\twocolumn[{%
\begin{center}

\vspace{0.2in}

{\fontsize{22}{26}\selectfont\sffamily\bfseries VISTA Architect: A graph database-oriented health AI system demonstrated in multidisciplinary tumor boards\par}

\vspace{0.7em}

{\sffamily\normalsize
Tuomo Kiiskinen$^{1,*}$,
Jason Fries$^{1}$,
Philip Adamson$^{1}$,
David Wu$^{2}$,
Timothy John Ellis-Caleo$^{2}$,
Aaron Fanous$^{1}$,
Balasubramanian Narasimhan$^{1}$,
Joel Neal$^{2}$,
Sylvia Plevritis$^{1}$,
Manuel A. Rivas$^{1,*}$
\par}

\vspace{0.4em}

{\sffamily\scriptsize\color{gray!70!black}
$^{1}$Department of Biomedical Data Science, Stanford University School of Medicine \;\;
$^{2}$Department of Medicine, Stanford University School of Medicine \\[0.2em]
$^{*}$Corresponding authors: {\fontfamily{ptm}\selectfont\href{mailto:tuomoki@stanford.edu}{tuomoki@stanford.edu}}, {\fontfamily{ptm}\selectfont\href{mailto:mrivas@stanford.edu}{mrivas@stanford.edu}}
\par}

\vspace{0.8em}

\parbox{0.92\textwidth}{%
\noindent\textbf{\sffamily Abstract}\\[0.3em]
\noindent\small
We introduce \textit{VISTA Architect}, a database-oriented AI architecture for
integrating large language models (LLMs) with longitudinal electronic health records
(EHRs). At ingestion, it transforms complex clinical documentation into a persistent,
provenance-linked knowledge graph, eliminating repeated reprocessing of raw records at
query time. The architecture has two layers: a source-faithful \emph{MEDS Graph}
preserving granular EHR structure with full provenance, and a clinically abstracted
\emph{Timeline Object Architecture} (TOA) that uses graph-guided LLM extraction to
synthesize a concise timeline of deduplicated, temporally coherent clinical events.
This addresses key limitations of direct long-context prompting and
retrieval-augmented generation (RAG), which often miss temporal relationships and
incur high cost and latency from repeated raw-text processing. By precomputing
clinical synthesis once, downstream queries access an organized patient state and
traverse to source documentation only when detailed verification is needed.

We demonstrate the system in multidisciplinary thoracic oncology tumor boards at
Stanford Medicine, where precise reconstruction of patient histories is critical.
Across 1{,}180 patients, VISTA Architect achieved 96.4\% accuracy (mean 9.75/10) on 15
tumor board--salient variables (17{,}700 evaluations; 95\% CI 96.1--96.7\%),
surpassing a matched BM25 RAG baseline and recent benchmarks for LLM-based clinical
extraction. An agentic interface reduced preparation for a 30-patient held-out cohort
to about 2.2~minutes without sacrificing accuracy. While configured here for thoracic
oncology, the modular design adapts to other specialties through customizable event
definitions, episode structures, and agentic tools; validation beyond thoracic
oncology remains future work.
}

\vspace{0.8em}
\rule{0.92\textwidth}{0.4pt}
\vspace{0.4em}

\end{center}
}]


\section{Introduction}

Clinicians routinely rely on electronic health records (EHR) to reconstruct complex patient narratives that guide critical clinical decisions. However, EHR data are primarily organized for documentation rather than for clinical reasoning, fragmenting patient trajectories across isolated records such as progress notes, laboratory results, and imaging reports.\cite{budd2023burnout} The burden of manually integrating these fragmented data into coherent clinical narratives falls heavily on clinicians, who often spend considerable time---up to 4~hours daily---on chart review alone,\cite{sinsky2016allocation} exacerbating burnout and increasing the risk of cognitive errors.\cite{budd2023burnout,han2019estimating}

Multidisciplinary tumor boards exemplify these challenges.\cite{hammer2020digital,nobori2022electronic,chang2025use} These real-time care conferences require rapid and precise synthesis of complex longitudinal patient trajectories---from initial diagnosis through treatments, responses, and disease progression---to inform critical medical decision-making without direct patient interaction. Thoracic oncology tumor boards, for example, rely heavily on accurate temporal narratives, including distinctions between baseline disease, treatment effects, emerging resistance mutations, and toxicity-related complications. Currently, preparing a single patient case for tumor board discussion demands approximately 15--30~minutes per clinician and suffers from significant inter-observer variability, potentially impacting the consistency and quality of clinical decisions.\cite{hammer2020digital,nobori2022electronic}

Traditional computational approaches attempting to address this challenge face a fundamental tension between the scale of longitudinal patient records and the capacity of current AI methods. For individual clinical documents, large language models (LLMs) have demonstrated strong summarization capabilities, in some cases matching or exceeding medical experts.\cite{van2024adapted,jiang2023health} However, tumor board preparation requires synthesis over longitudinal documentation that may span years of care, thousands of clinical events, and very large text volumes. This scale often exceeds the practical context available to current LLM workflows, particularly when the same patient record must be queried repeatedly across multiple clinical questions. When applied to such longitudinal records, LLM performance degrades: temporal reasoning deteriorates as input length increases,\cite{cui2025timer} and hallucinations and omissions become clinically consequential.\cite{asgari2025framework} Retrieval-augmented generation (RAG) offers a strategy for navigating records beyond context length,\cite{lewis2020retrieval} but retrieves fragments by semantic similarity rather than temporal or clinical structure, producing disconnected context that may miss critical longitudinal relationships. Dense retrieval models frequently underperform simple lexical baselines in specialized domains,\cite{thakur2021beir} and even optimized clinical RAG pipelines require substantially more computation per query while achieving suboptimal accuracy relative to structured approaches.\cite{lopez2025clinical} Empirically, reported accuracy levels for LLM-based clinical information extraction reflect these limitations: a recent meta-analysis of 56 oncology studies reported an average overall accuracy of 76.2\% (diagnostic accuracy 67.4\%),\cite{hao2025large} agentic multi-agent orchestration achieved 84\% strict recall on tumor board--salient facts across 71 patients,\cite{blondeel2025healthcare} and optimized clinical RAG pipelines for note-level extraction reported F1 scores of 0.79--0.90 depending on retrieval strategy.\cite{lopez2025clinical} These results were typically obtained on note-level or short-document tasks; whole-record longitudinal extraction is harder still. Relational databases can support deterministic queries and provenance, but reconstructing a patient-level clinical state from normalized EHR tables remains a domain-specific synthesis problem: events must be temporally aligned, repeated documentation must be reconciled, and clinically meaningful phases must be inferred. VISTA Architect addresses this by making longitudinal synthesis explicit and amortized at ingestion rather than re-derived for each downstream application.\cite{hripcsak2015observational}

Here we introduce \textit{VISTA Architect}, a database-oriented AI system that addresses a general challenge in applying LLMs to large document repositories: how to provide accurate, structured, and complete context from records that far exceed model capacity, without sacrificing retrieval speed or temporal fidelity.\cite{shang2024electronic} Our approach employs a two-tiered graph architecture that pre-computes a structured representation of the available longitudinal clinical record. The first tier, a hierarchical \emph{MEDS Graph}, preserves the granular documentation structure of the source data. The second tier, a clinically focused \emph{Timeline Object Architecture} (TOA), synthesizes these into deduplicated, temporally ordered events and episodes---producing a queryable knowledge graph that any downstream application can traverse deterministically. Because the clinically organized graph is constructed once and queried repeatedly, the expensive part of the workflow---longitudinal synthesis into events, episodes, and current-state artifacts---is amortized. At query time, applications operate over this precomputed clinical context and traverse provenance links only when targeted source evidence is needed, rather than repeatedly searching or rereading the raw record. The contribution is therefore not that a graph store categorically replaces SQL, but that VISTA encodes a clinically meaningful, provenance-linked hierarchical tree-like compression of the longitudinal record, allowing downstream agents to navigate from patient state to targeted source evidence rather than reconstructing clinical context from raw records at query time. On top of this persistent graph layer, every interaction with VISTA Architect---patient build, dashboard population, chat, and cross-patient retrieval---is served by an agentic interface in which the receiving component plans its own graph traversals and parallel sub-agent invocations, supporting both on-demand cohort preparation and conversational clinical reasoning from a single shared graph. We demonstrate this architecture in the context of electronic health records (EHR) and multidisciplinary tumor boards, where the demands on accuracy, temporal precision, and speed are particularly acute.

At our own institution, a manual LLM-based pre-tumor-board summarization workflow has been developed, evaluated against physician-authored references, and deployed into routine thoracic tumor board practice.\cite{elliscaleo2026mtb} VISTA Architect builds on this operational foundation: it pre-computes a complete structured patient timeline, from which the same clinician-facing summary of critical information is retrieved automatically rather than summarized from a fixed window of recent notes for each query.

We validate the VISTA Architect system in the highly demanding context of thoracic
oncology tumor boards at Stanford Medicine. Across the full cohort of 1{,}180
patients, VISTA Architect achieved 96.4\% accuracy (mean score 9.75/10; 95\% CI
96.1--96.7\%) on 15 tumor board--salient clinical variables, with 17{,}063 of 17{,}700
variable evaluations rated as Correct. Automated case preparation completed in 2.2 minutes for a held-out 30-patient cohort under the fully agentic build, matching and extending recent efforts exploring agentic LLM orchestration for patient summarization in tumor-board workflows.\cite{blondeel2025healthcare}

Through its structured yet flexible design, VISTA Architect offers a model for connecting large document repositories to LLM-powered applications. The architecture is adaptable via specialty-specific configurations---defining context-appropriate node types, episode structures, and variable dictionaries---making it applicable beyond thoracic oncology to other clinical specialties and, more broadly, to any domain where large, longitudinal document collections must be queried with high accuracy and complete provenance.

\section{Results}

\subsection{Architecture}

\begin{figure*}[!t]
    \centering
    \includegraphics[width=\linewidth]{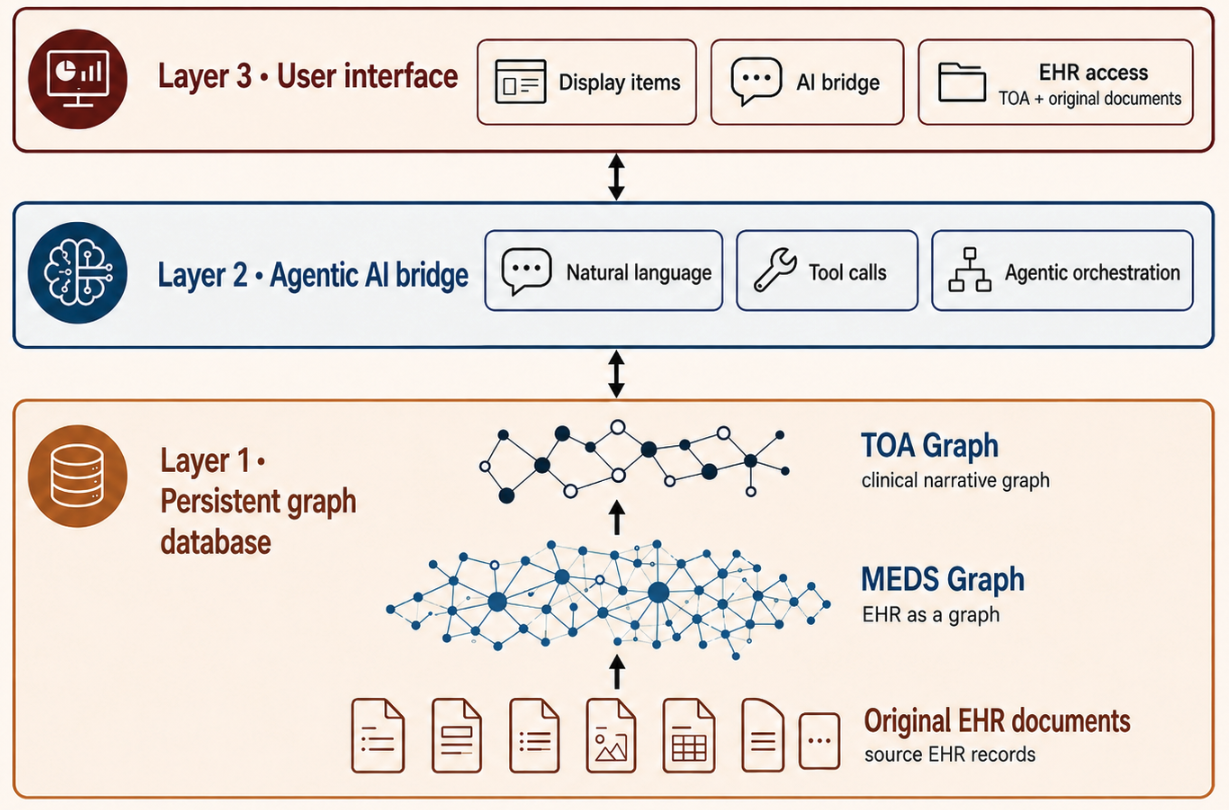}
    \caption{\textbf{VISTA Architect high-level system design.} VISTA Architect converts source EHR data into a persistent graph database, consisting of a source-faithful MEDS Graph and a clinically abstracted TOA Graph. The agentic AI bridge operates over TOA events, episodes, current-state artifacts, and targeted MEDS Graph retrieval through tool calls. The user interface exposes the same graph-resident artifacts through display items, natural-language interaction, and provenance-linked EHR access.}
    \label{fig:system_architecture}
\end{figure*}

We designed VISTA Architect using a database-oriented AI framework, diverging from text-processing methodologies such as direct LLM prompting, retrieval-augmented generation (RAG) over raw EHR text, and programmatic deterministic queries over relational EHR schemas, by positioning graph databases at the core of clinical data representation.\cite{shang2024electronic} The architecture is organized into three interconnected layers: (1) a structured multi-layered graph database derived directly from electronic health record (EHR) data; (2) an AI bridge that manages complex query translation and data retrieval; and (3) a generative and agentic user interface that dynamically generates display items directly from the TOA data layer.

The graph database is constructed in three sequential steps during patient ingestion, each producing a graph-resident layer that downstream applications can query deterministically. The two subsections that follow describe how these layers are assembled into the runtime context for an agentic AI bridge, and how that bridge is exposed through the user-facing applications.

\subsubsection{Step 1: Turning the EHR into a graph database}

First, VISTA Architect transforms the source health record into a persistent graph database. In the implementation evaluated here, the source data were represented as MEDS XML, a hierarchical representation derived from the Medical Event Data Standard (MEDS).\cite{arnrich2024medical} However, the architecture does not depend on MEDS XML specifically. The same graph layer can be constructed from other structured, semi-structured, or document-based clinical sources, provided that the source data can be represented as nodes, edges, timestamps, and provenance links.

This first graph layer, the MEDS Graph, represents the EHR as a graph. It preserves the granular structure of the source record: patients, encounters, notes, measurements, procedures, medications, imaging records, and other clinical entries become graph nodes, while temporal, visit-level, and source-document relationships become graph edges. The goal of this layer is not to summarize the record, but to make the original EHR computationally addressable.

This step is deterministic and does not require an LLM. Because the MEDS Graph maintains links back to the original source documents, later AI-generated assertions can be traced to the underlying EHR evidence. The graph can also link to multimodal assets such as imaging studies, audio recordings, and waveform data without storing them directly, keeping the graph lightweight to operate. The graph database therefore becomes the primary data structure of the system: the LLM is not asked to repeatedly reread the full record from scratch, but instead operates over a structured representation of the record. In our thoracic oncology cohort, per-patient MEDS Graphs contained a median of 3{,}608 nodes (IQR 562--11{,}629; the full MEDS Graph schema is given in Appendix~\ref{app:meds_schema}).

\subsubsection{Step 2: Timeline Object Architecture (TOA)}

Next, VISTA Architect constructs a compact clinical timeline graph on top of the source-level graph. We refer to this layer as the Timeline Object Architecture (TOA) Graph. Whereas the MEDS Graph represents the EHR as documented, the TOA Graph represents the patient trajectory as a coherent timeline of clinically meaningful events; both graph layers are shown for a single representative patient in Supplementary Fig.~\ref{fig:graph_structure}.

The central operation in this step is transforming EHR data from documents into clinical events. For example, a diagnosis, surgery, imaging result, molecular test, treatment start, adverse event, or change in clinical status may be mentioned across multiple notes. In the TOA layer, these mentions are resolved into a single event object linked back to the source evidence in the MEDS Graph. This converts the longitudinal record from a documentation-centered graph into a decision-centered clinical timeline.

This step uses graph-guided LLM extraction. Rather than prompting an LLM with an entire raw EHR, the system traverses the MEDS Graph, splits the EHR into compact graph-derived segments, presents those segments to the model, and asks it to produce standardized, structured clinical event objects. The LLM therefore performs a bounded abstraction task over graph-derived context, while the graph preserves provenance and temporal structure. A key distinction is that timestamps in the TOA Graph represent the inferred time of the actual clinical occurrence, whereas timestamps in the source EHR typically reflect when that information was documented. In some cases these dates are effectively identical; in others, the difference may span months, years, or even decades when historical events are recorded retrospectively. The TOA Graph therefore represents the patient's actual biomedical and clinical course rather than the order in which facts happened to appear in notes. The event-extraction and episode-synthesis prompts used in the thoracic oncology configuration are reproduced in Appendices~\ref{app:event_extraction_prompt} and~\ref{app:episode_synthesis_prompt}.

\subsubsection{Step 3: Forming episodes and current-state artifacts}

Finally, the TOA graph is expanded by generating higher-order clinical structures from the timeline object event stream. Individual TOA events provide a detailed timeline, but episode formation compresses that stream into a compact clinical story: meaningful phases with clear transitions, anchors, and reasons for change. In the thoracic oncology configuration, these episodes include baseline history, diagnostic workup, treatment lines, and post-oncological phases, but the episode vocabulary is configurable for other domains.

Episode formation serves two purposes. First, it makes complex longitudinal records easier for humans to understand by organizing long sequences of events into clinically meaningful phases. Second, it is itself a precomputed reasoning step that aids later AI operations over the graph. By storing phase boundaries, transition points, treatment contexts, and termination reasons, the system gives downstream agents a structured map of the timeline rather than only a flat sequence of events. Later graph operations can therefore start from a clinically organized representation and traverse into finer-grained evidence only when needed.

The same ingestion process also produces current-state graph-resident artifacts, such as structured variable snapshots, summaries, and display-ready fields. These artifacts represent the system's latest synthesized view of the patient at the relevant decision point. They are not merely UI elements; they are stored outputs of prior reasoning over the graph. Dashboards, agents, documentation tools, cohort queries, and cross-patient retrieval can reuse these artifacts without reconstructing the clinical context from the raw EHR for every interaction.

\subsubsection{Connecting the graph to an agentic AI bridge}

At runtime, the AI bridge operates with the TOA layer, episode structure, current-state artifacts, and a small recent EHR context available to the model. This gives the agent a compact but clinically organized view of the patient before any additional retrieval is needed. User or application requests can therefore often be answered directly from the graph-resident clinical representation.

When additional detail is required, the agent uses the TOA layer to navigate the lower-level MEDS Graph and retrieve targeted pieces of the original EHR. For example, the TOA Graph may identify the relevant clinical event or episode, after which the agent can traverse provenance edges to recover the exact source note, imaging report, laboratory value, or document fragment. Retrieval is therefore guided by the clinical abstraction layer rather than performed as an unguided search over raw text. In this sense, graph traversal is used as a clinical navigation operation: the agent starts from TOA events, episodes, or current-state artifacts and follows clinically meaningful edges to the relevant source-level nodes, rather than issuing a broad search over the full patient record.

This design changes the role of retrieval. Conventional RAG systems retrieve fragments from raw documents at query time. VISTA Architect instead performs much of the semantic and temporal organization once, stores it as graph structure, and then uses that structure to retrieve only the targeted source evidence needed for a given task. The AI bridge uses natural language, tool calls, and agentic orchestration, but the durable context remains the graph: TOA for clinical navigation, MEDS Graph for source-level detail, and original EHR documents for auditability.

\subsubsection{Exposing the architecture through user-facing applications}

The final step is the user-facing application layer. Users see many of the same display artifacts that the AI bridge holds in context: structured patient fields, timeline summaries, episode-level views, and current-state summaries. This makes the interface and the AI bridge operate over the same underlying representation rather than over separate data products. In the thoracic oncology configuration, this layer exposes Overview, Clinical Timeline, automated tumor board note, and chat views, all rendered from graph-resident artifacts (Supplementary Figs.~\ref{fig:ui_overview}, \ref{fig:ui_timeline}, \ref{fig:ui_note}, and~\ref{fig:ui_chat_interface}).

From the user perspective, this enables both direct review and interactive extension. A user can inspect display items, ask questions about them through the AI bridge, and, when needed, the agent can perform additional graph operations. These operations can be purely retrieval-oriented, such as finding the source evidence for a clinical assertion, or generative, such as drafting a note, producing a summary, creating a plot, or assembling a domain-specific view.

Here we demonstrate VISTA Architect in multidisciplinary thoracic tumor boards. The tumor-board application is a configuration of the system, not the boundary of the architecture. The architecture can be adapted to other clinical domains through configurable event types, episode structures, display artifacts, edge types, and agentic tool capabilities. What is shared is the architectural core: construct a source-faithful EHR graph, abstract it into a clinically meaningful TOA graph, store reusable reasoning artifacts, and expose the result through an agentic AI bridge and user interface.

\subsection{Application to Thoracic Tumor Boards}

\subsubsection{Clinical Context}

Multidisciplinary tumor boards require clinicians to rapidly synthesize comprehensive patient histories from fragmented and complex electronic health record (EHR) data.\cite{hammer2020digital,nobori2022electronic,chang2025use} These real-time care conferences integrate expertise from radiologists, pathologists, medical oncologists, surgeons, and geneticists to develop precise, patient-specific treatment strategies. Accurate and temporally precise reconstruction of clinical events---including diagnostic evaluations, therapeutic interventions, molecular testing, and toxicity management---is essential to informed clinical decision-making in oncology, particularly in thoracic malignancies where patients frequently experience multiple treatment stages and complex care trajectories.

\begin{figure*}[!t]
  \centering
  \includegraphics[width=0.95\textwidth]{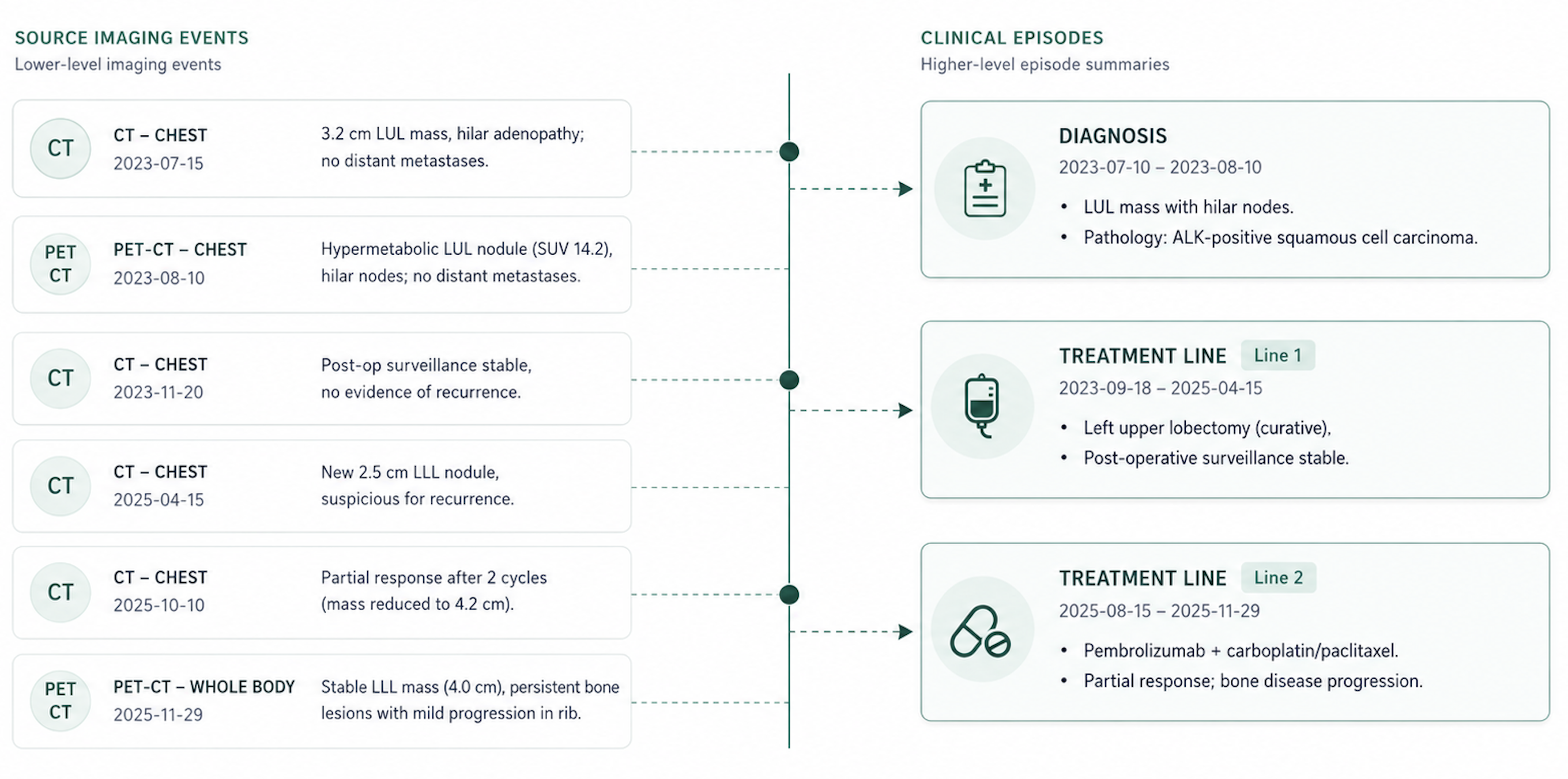}
  \caption{\textbf{Synthetic example of TOA-based timeline organization.} VISTA Architect maps source-level imaging events from the EHR into graph-resident clinical episodes. The central timeline shows how granular events contribute to higher-order episode summaries, preserving temporal structure and provenance while creating a compact representation for display, agentic retrieval, and source-evidence navigation.}
  \label{fig:timeline_example}
\end{figure*}

\subsubsection{Implementation Details}

We implemented VISTA Architect specifically within thoracic oncology tumor boards, leveraging its flexible architecture through configurable profiles and specialized prompts. These profiles define how the MEDS Graph and Timeline Object Architecture (TOA) layers process and structure clinical data specifically for thoracic oncology decision-making.

For thoracic oncology tumor boards, the profile configuration involved defining clinically meaningful episodes and events tailored explicitly for oncology scenarios. Episodes included baseline assessments, diagnostic evaluations, distinct numbered treatment lines, and post-treatment surveillance. Imaging events such as CT, MRI, and PET scans served as anchors for defining these episodes, with events grouped around these anchors within clinically meaningful temporal windows ($\pm$90 days overall, $\pm$7 days for co-occurrence).

Additionally, we established a focused variable dictionary capturing the most clinically relevant patient information frequently discussed during tumor board meetings. These variables included demographic details (date of birth, sex, smoking status), tumor-specific factors (diagnosis, histology, metastasis, lymph node involvement, and genetic testing panel), treatment information (previous surgeries, current medical therapy, radiation therapy, last imaging dates), clinical status (ECOG performance status, therapy toxicity/comorbidities), and safety-critical details (allergies). The full dictionary, with variable types, is provided in Supplementary Table~\ref{tab:variables}.

The variable dictionary and the pre-tumor-board summary note were not defined de novo: they follow the tumor-board--salient fields and the concise summary format already in operational use at the Stanford thoracic tumor board.\cite{elliscaleo2026mtb} VISTA Architect produces these same artifacts from the precomputed graph at ingestion, rather than re-deriving them from per-query note retrieval.

\subsection{Clinical Accuracy}

The 1{,}180 patients in this study were those in the Data Lake who presented at
thoracic multidisciplinary tumor boards between (de-identified) January 2020
and November 2025 (Table~\ref{tab:cohort}). The cohort was predominantly NSCLC
(82.0\%), with representation of thymoma/thymic tumors (8.0\%), SCLC (2.9\%),
mesothelioma (2.9\%), and other thoracic malignancies (4.2\%). Molecular profiling
results were documented in 67.3\% of patient records, with EGFR (17.8\%),
KRAS (8.2\%), and ALK (4.1\%) the most prevalent actionable drivers across the
full cohort. Each patient contributed an EHR data export, converted to MEDS
XML (Section~2.1), with a median record time span of 8.5 years (IQR 4.4--12.7), a
median of 3{,}255 clinical events (IQR 362--10{,}922), and a median of 3{,}608
MEDS Graph nodes (IQR 562--11{,}629), encompassing demographics, oncologic
procedures, and longitudinal clinical text. From each patient, we extracted 15
multidisciplinary tumor board (MTB)-salient variables spanning demographics, tumor
characteristics, treatment history, clinical status, imaging, and safety
(Table~\ref{tab:per_variable}), yielding 17{,}700 total variable evaluations.

\begin{table*}[!t]
\centering
\small
\setlength{\tabcolsep}{4pt}
\caption{Cohort characteristics (N=1{,}180), summarized from VISTA-generated structured artifacts. Values combine source-structured fields and text-derived artifacts produced during pipeline processing; this descriptive table is not intended as a standalone benchmark of LLM extraction from unstructured text. Per-gene rows report positive results; patients with multiple actionable drivers are counted in more than one row, so the per-gene counts sum to more than the 613 patients with any actionable driver.}
\label{tab:cohort}
\begin{tabular}{@{}p{7.2cm} r@{}}
\toprule
\textbf{Characteristic} & \textbf{Value} \\
\midrule
\multicolumn{2}{@{}l}{\textbf{Patient Demographics}} \\
\quad Age at tumor board, median (IQR), y & 69 (61--76) \\
\quad Male & 547 (46.4\%) \\
\quad Female & 633 (53.6\%) \\
\midrule
\multicolumn{2}{@{}l}{\textbf{Clinical Characteristics}} \\
Smoking status \\
\quad Current / Former / Never & 74 / 470 / 575 \\
\quad Unknown & 61 (5.2\%) \\
\quad Pack-years, median (IQR) & 30 (12--45) \\
ECOG performance status \\
\quad 0 / 1 / 2 / 3--4 & 254 / 379 / 76 / 29 \\
\quad Unknown & 343 (29.1\%) \\
\midrule
\multicolumn{2}{@{}l}{\textbf{Tumor Characteristics}} \\
Primary diagnosis \\
\quad NSCLC & 968 (82.0\%) \\
\quad Thymoma/thymic tumors & 94 (8.0\%) \\
\quad SCLC & 34 (2.9\%) \\
\quad Mesothelioma & 34 (2.9\%) \\
\quad Other thoracic malignancies & 50 (4.2\%) \\
Metastatic disease \\
\quad Yes / No & 523 / 500 \\
\quad Unknown or suspected & 157 (13.3\%) \\
Lymph node involvement \\
\quad Yes / No & 572 / 437 \\
\quad Unknown or suspected & 171 (14.5\%) \\
\midrule
\multicolumn{2}{@{}l}{\textbf{Molecular Testing}} \\
\quad Results documented & 794 (67.3\%) \\
\quad Any actionable driver mutation & 613 (51.9\%) \\
\quad\quad EGFR / KRAS / ALK & 210 / 97 / 48 \\
\quad\quad MET / HER2 / BRAF & 35 / 29 / 26 \\
\quad\quad ROS1 / RET / NTRK & 19 / 19 / 5 \\
\quad PD-L1 results documented & 671 (56.9\%) \\
\midrule
\multicolumn{2}{@{}l}{\textbf{Treatment History}} \\
\quad Current systemic anticancer therapy & 385 (32.6\%) \\
\quad History of radiation therapy & 389 (33.0\%) \\
\midrule
\multicolumn{2}{@{}l}{\textbf{Safety}} \\
\quad Known drug allergies & 545 (46.2\%) \\
\bottomrule
\end{tabular}
\end{table*}
We evaluated VISTA Architect across the full 1{,}180-patient thoracic oncology cohort
using 15 tumor board--salient variables per patient (17{,}700 total evaluations).
Ground truth was established from raw EHR XML truncated at the documented tumor board
date, ensuring evaluation reflects only information available at the clinical decision
 point. The production pipeline used GPT-4.1 for chunk-level event extraction and GPT-5
for downstream patient-info synthesis; an independent GPT-5 LLM-as-judge (Appendix~\ref{app:judge_prompt}) scored each
extracted variable against the ground truth, with full evaluation protocol and clinician
validation described in Methods.

Overall accuracy was high, achieving an approximate mean quality score of \textbf{9.75/10} with \textbf{17{,}063/17{,}700 (96.4\%; 95\% CI 96.1--96.7\%)} variables rated as Correct (Table~\ref{tab:aggregate}).\footnote{DNR / code status, evaluated in the original 16-variable schema and excluded from the primary endpoint because it is deterministically available from OMOP-coded fields, was extracted at mean 9.92/10 and 99.5\% correct (1{,}174/1{,}180; Wilson 95\% CI 98.9--99.8\%).} As a sensitivity analysis addressing the concern that two of the 15 reported variables (Date of Birth and Sex) are reliably available from structured OMOP fields and serve here as positive controls for upstream hallucination (Section~4.2), the same headline excluding those two positive controls is 9.71/10 mean (95.85\% correct; 14{,}703/15{,}340 evaluations across 13 variables, Wilson 95\% CI 95.52--96.15\%).

\begin{table*}[!t]
\centering
\begin{tabular}{@{}ll@{}}
\toprule
\textbf{Metric} & \textbf{Value} \\
\midrule
Patients evaluated & 1{,}180 \\
Variables per patient & 15 \\
Total evaluations & 17{,}700 \\
Mean quality score & \textbf{9.75/10} \\
Correct & \textbf{17{,}063/17{,}700 (96.4\%; 95\% CI 96.1--96.7\%)} \\
Incorrect & 637/17{,}700 (3.6\%) \\
\bottomrule
\end{tabular}
\caption{Aggregate extraction accuracy across 15 MTB-salient variables for 1{,}180
patients after excluding DNR/code status from the primary endpoint.}
\label{tab:aggregate}
\end{table*}

Performance was consistent across all six clinical categories
(Table~\ref{tab:per_variable}). Demographics achieved the highest accuracy: Date of
Birth and Sex were both extracted perfectly (10.00/10, 100\% correct across all
1{,}180 patients; 95\% CI 99.7--100\%). Safety- and treatment-related variables were extracted with high reliability, including Radiation Therapy at 9.94/10 (99.1\%; 95\% CI 98.3--99.5\%) and Allergies at 9.70/10 (94.8\%; 95\% CI 93.4--96.0\%). Tumor characterization variables
were reliably extracted, with Histology at 9.89/10 (97.8\%), Genetic Testing Panel at
9.75/10 (95.4\%), and both Metastasis and Lymph Node Involvement at 9.68--9.69/10
(94.8--95.0\%). The most challenging variable was Date of Last CT (9.52/10, 92.1\%;
95\% CI 90.4--93.5\%), reflecting ambiguity between internal and external imaging
dates.

\begin{table*}[!t]
\centering
\small
\begin{tabular}{@{}llcccc@{}}
\toprule
\textbf{Variable} & \textbf{Category} & \textbf{Mean Score} & \textbf{95\% CI} &
\textbf{\% Correct} & \textbf{95\% CI} \\
\midrule
Date of Birth & Demographics & 10.00 & (10.00--10.00) & 100.0\% & (99.7--100.0\%) \\
Sex & Demographics & 10.00 & (10.00--10.00) & 100.0\% & (99.7--100.0\%) \\
Smoking Status & Demographics & 9.80 & (9.75--9.84) & 97.8\% & (96.8--98.5\%) \\
\midrule
Diagnosis & Tumor & 9.72 & (9.68--9.77) & 95.3\% & (94.0--96.4\%) \\
Histology & Tumor & 9.89 & (9.86--9.93) & 97.8\% & (96.8--98.5\%) \\
Metastasis & Tumor & 9.68 & (9.62--9.74) & 95.0\% & (93.6--96.1\%) \\
Lymph Node Involvement & Tumor & 9.69 & (9.63--9.75) & 94.8\% & (93.4--96.0\%) \\
Genetic Testing Panel & Tumor & 9.75 & (9.69--9.80) & 95.4\% & (94.1--96.5\%) \\
\midrule
ECOG Performance Status & Clinical & 9.69 & (9.64--9.73) & 98.1\% & (97.2--98.8\%) \\
Therapy Toxicity / Comorbidities & Clinical & 9.58 & (9.54--9.62) & 95.0\% &
(93.6--96.1\%) \\
\midrule
Previous Surgery & Treatment & 9.53 & (9.44--9.63) & 95.2\% & (93.8--96.3\%) \\
Current Medical Therapy & Treatment & 9.72 & (9.66--9.79) & 95.5\% & (94.2--96.5\%) \\
Radiation Therapy & Treatment & 9.94 & (9.91--9.96) & 99.1\% & (98.3--99.5\%) \\
\midrule
Date of Last CT & Imaging & 9.52 & (9.43--9.62) & 92.1\% & (90.4--93.5\%) \\
\midrule
Allergies & Safety & 9.70 & (9.63--9.77) & 94.8\% & (93.4--96.0\%) \\
\midrule
\textbf{Overall} & \textbf{All} & \textbf{9.75} & \textbf{--} &
\textbf{96.4\%} & \textbf{(96.1--96.7\%)} \\
\bottomrule
\end{tabular}
\caption{Per-variable extraction accuracy with 95\% confidence intervals after
excluding DNR/code status from the primary endpoint (N=1{,}180 patients, 17{,}700
evaluations). Mean scores on 0--10 scale. CIs for mean scores via normal
approximation; CIs for \% correct via Wilson score interval.}
\label{tab:per_variable}
\end{table*}

Independent clinician validation of the LLM-as-judge was performed on a randomly selected 30-patient subset from the full 1{,}180-patient cohort. The original clinician review covered 480 paired evaluations from the initial 16-variable set; the primary endpoint reported here excludes DNR/code status, yielding 450 endpoint evaluations in the same 30 patients. The subset was subsequently checked to ensure that it was representative of the broader cohort rather than an outlier sample. Clinician review classified each of the 480 paired judge decisions as Agree, Uncertain, or Disagree: 95.4\% were rated Agree (95\% CI 93.2--97.0\%), 4.6\% Uncertain, and 0\% Disagree (zero of 480; 95\% CI upper bound 0.79\%). The Uncertain cases clustered on variables with inherent clinical ambiguity. The judge correctly
identified all six VISTA extraction failures in the sample (100\% sensitivity),
grading each as Partial or Incorrect.

On the same 30-patient cohort, we additionally evaluated the fully agentic
extraction pipeline (Section~2.1.4), in which chunk-level extraction is
parallelized within and across patients and coordinated by an agent. This agentic
build also used a different underlying model family from the sequential
pipeline---Claude Opus 4.6 as orchestrator, with Gemini 3.5 Flash for per-chunk
event extraction and episode synthesis (Appendix~\ref{app:agentic_implementation})---so
the agreement reported here additionally indicates that extraction quality is not
tied to a single model provider. Extraction
accuracy remained consistent with the sequential pipeline reported above: the
GPT-5 judge scored the two pipelines within confidence-interval overlap on the
16-variable rubric, with no systematic per-variable regression. The wall-time
gains achieved under this configuration are reported in Section~2.5.

Error analysis across the 637 incorrect extractions (3.6\%) revealed systematic
patterns rather than random failures. The most common error source was imaging
chronology (Date of Last CT), arising from ambiguity between internal and external
imaging dates or PET-CT versus standalone CT attribution. Previous Surgery errors
stemmed from boundary cases in surgical procedure classification (e.g., diagnostic
VATS biopsy vs.\ oncologic resection). Allergies and Current Medical Therapy errors
reflected documentation fragmentation across the longitudinal record. Metastasis
errors arose from ambiguous staging scenarios where imaging was ``suspicious but not
confirmed.'' No errors occurred in Date of Birth or Sex across all 1{,}180 patients. Worked examples of a perfect extraction and of a representative imaging-chronology failure mode are provided in Supplementary Tables~\ref{tab:synth_perfect} and~\ref{tab:synth_failure}.

\subsection{Comparison with RAG Baseline}
\label{subsec:rag_comparison}

Additionally, we performed the same retrieval task using a standard BM25
retrieval-augmented generation (RAG) baseline on the randomly selected,
representative 30-patient subset (450 total evaluations after excluding DNR/code
status from the primary endpoint), using the same LLM-as-judge protocol and
identical ground truth (Appendix~\ref{app:rag_baseline}). The RAG baseline
employed BM25 retrieval over raw EHR XML entry-level chunks with focused,
per-variable queries (Appendix~\ref{app:rag_prompts})---the configuration most favorable to lexical retrieval---and
used the same generation models (GPT-4.1 and GPT-5) as VISTA Architect. No graph
structure, event extraction, or pre-computed artifacts were provided to the RAG
system.

After excluding DNR/code status from the primary endpoint, VISTA Architect achieved
96.9\% accuracy (mean 9.76/10) on this subset, consistent with the full-cohort
result. The RAG baselines achieved 66.7--66.9\% accuracy (mean approximately
7.58--7.73/10), a roughly 30 percentage-point gap in strict correctness
(Table~\ref{tab:rag_comparison}; full per-variable breakdown in Supplementary
Table~\ref{tab:rag_per_variable}). VISTA outperformed RAG on every one of the 30
patients (all paired deltas positive). The paired superiority is statistically
significant by both an exact sign test (30/30 patients, two-sided $p = 1.86 \times
10^{-9}$) and a Wilcoxon signed-rank test (one-sided, VISTA $>$ RAG: $W = 465$,
$p = 8.6 \times 10^{-7}$ against GPT-5 RAG; $p = 8.7 \times 10^{-7}$ against
GPT-4.1 RAG); mean per-patient $\Delta$ score (VISTA $-$ RAG) was $+2.06$ for
GPT-5 RAG and $+2.21$ for GPT-4.1 RAG. The RAG
baseline performed comparably to VISTA on single-fact, lexically accessible
variables (Sex, Smoking Status, Histology) and dropped sharply on variables
whose values depend on temporal resolution of multiple mentions across the
longitudinal record---Previous Surgery (36.7\% vs.\ 96.7\%), Metastasis
(50.0--53.3\% vs.\ 96.7\%), and Radiation Therapy (33.3--73.3\% vs.\ 100\%). In
contrast, this pattern of underperformance on time-varying variables was not
observed for VISTA Architect, which scored $\geq$96.7\% on all three.

\begin{table*}[!t]
\centering
\small
\begin{tabular}{@{}lcccc@{}}
\toprule
\textbf{Category} & \textbf{VISTA} & \textbf{RAG (GPT-4.1)} & \textbf{RAG (GPT-5)} \\
\midrule
Demographics & 9.96 (100.0\%) & 8.26 (81.1\%) & 8.34 (80.0\%) \\
Tumor & 9.72 (96.0\%) & 7.97 (70.0\%) & 8.04 (70.0\%) \\
Clinical & 9.65 (96.7\%) & 6.10 (46.7\%) & 7.92 (68.3\%) \\
Treatment & 9.72 (96.7\%) & 7.26 (60.0\%) & 6.83 (52.2\%) \\
Imaging & 9.77 (96.7\%) & 7.17 (66.7\%) & 6.93 (63.3\%) \\
Safety & 9.67 (93.3\%) & 8.00 (66.7\%) & 7.43 (56.7\%) \\
\midrule
\textbf{Overall} & \textbf{9.76 (96.9\%)} & \textbf{7.58 (66.7\%)} & \textbf{7.73 (66.9\%)} \\
\bottomrule
\end{tabular}
\caption{VISTA Architect vs.\ RAG baseline accuracy by clinical category after
excluding DNR/code status from the primary endpoint (N=30 patients, 450 evaluations
per system). RAG uses BM25 retrieval over raw EHR XML with focused per-variable
queries---the configuration most favorable to lexical retrieval. Full per-variable
results in Appendix~\ref{app:rag_baseline}.}
\label{tab:rag_comparison}
\end{table*}

\subsection{System Performance}

The thoracic oncology tumor board experiments reported throughout this work used
across-patient parallelization with sequential chunk processing within each
patient. Under this regime (10 patients in parallel), the 30-patient cohort
processes in 10 minutes (Table~\ref{tab:parallel}). The sequential baseline (no
parallelization) required 74 minutes total processing time for the same
30-patient cohort.

Full intra- and inter-patient parallelization of the build pipeline is only
practical when the unifier stage can correctly reconcile events that appear in
multiple chunks; the agentic interface (Section~2.1.4) provides this reconciliation
and was used to measure end-to-end wall time under full parallelization. Running
the agentic build on the held-out 30-patient test cohort completed in
\textbf{2.2 minutes} end-to-end (mean per-patient build wall time 88~s, median
87~s, max 131~s for the bottleneck patient at 16 chunks and 158 TOA events).
Per-patient stage timings under the agentic pipeline were dominated by the
final display-synthesis step, with per-chunk extraction running concurrently in
$\sim$21~s on average and the unifier and episode-synthesis steps each adding
under a few seconds. Extraction quality under the agentic pipeline on the
test30 cohort is reported in Section~2.3; the exact per-stage model assignments and
orchestration are detailed in Appendix~\ref{app:agentic_implementation}.

The real-world workflow speedup from preparation automation remains to be determined prospectively. As a reference point, establishing chart-reviewed ground truth for the evaluation required approximately 40--60 minutes per patient for a clinician to review and annotate the target variables. This places manual ground-truth construction for the 30-patient validation subset at roughly 20--30 clinician-hours, compared with 2.2 minutes of automated processing under the fully agentic build.

\begin{figure*}[!t]
    \centering
    \includegraphics[width=\linewidth]{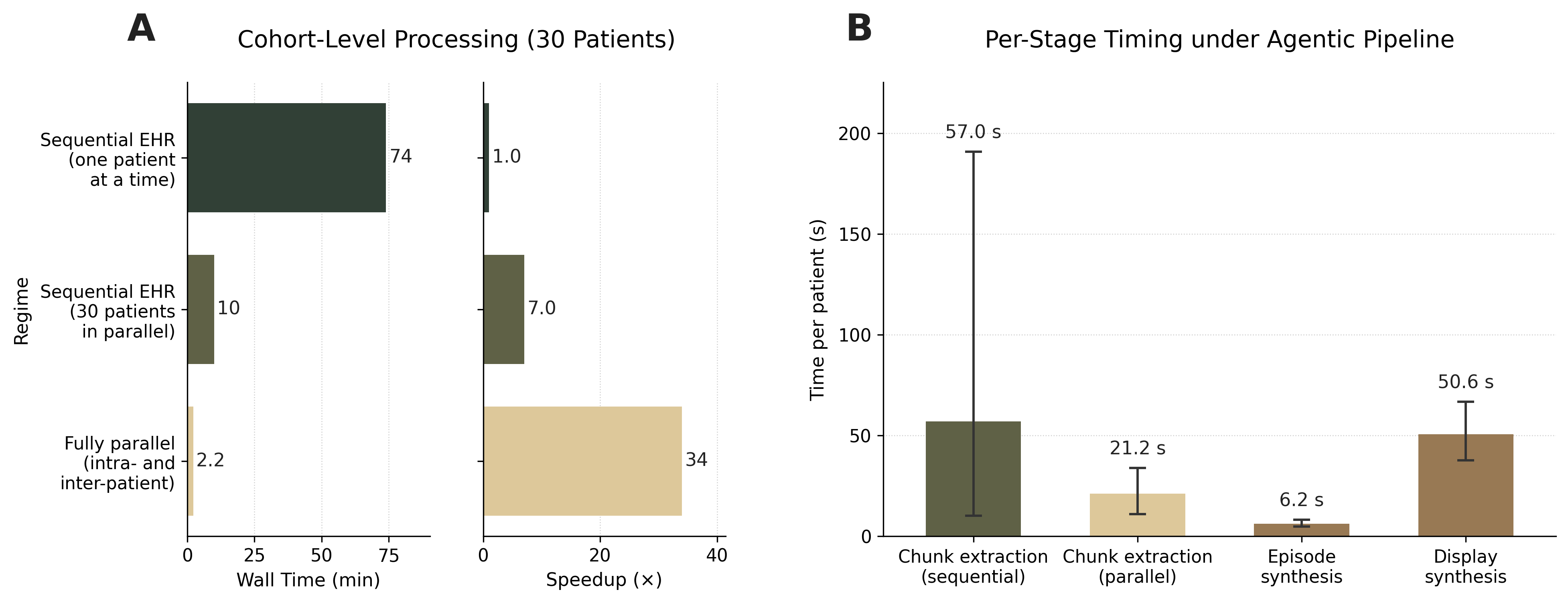}
    \caption{Single-patient MEDS Graph construction performance for VISTA Architect.
    Build time and peak memory usage are shown for the programmatic XML-to-graph
    conversion step for representative small
    ($<$1\,MB XML, $\sim$500 MEDS Graph nodes), mid-sized ($\sim$2\,MB XML, $\sim$2{,}200 nodes),
    and large ($\sim$5\,MB XML, $\sim$10{,}000 nodes) patients.
    All MEDS Graphs are constructed in under one second and under 20\,MB of memory,
    showing approximately linear scaling with patient record size. Note that
    the subsequent LLM-based TOA event extraction step requires additional time;
    per-patient end-to-end build times under the fully agentic pipeline are
    reported in Section~2.5 (mean 88\,s, median 87\,s, max 131\,s on the
    test30 cohort).}
    \label{fig:build_performance}
\end{figure*}

\begin{table*}[!t]
\centering
\begin{tabular}{lcc}
\toprule
\textbf{Regime} & \textbf{Wall Time} & \textbf{Speedup} \\
\midrule
Sequential (one patient at a time, sequential chunks) & 74 min  & 1.0$\times$ \\
Across-patient (10 patients in parallel, sequential chunks within patient) & 10 min & 7$\times$ \\
\textbf{Fully agentic (intra- and inter-patient parallel, measured)} & \textbf{2.2 min} & \textbf{34$\times$} \\
\bottomrule
\end{tabular}
\caption{Parallelization impact for the 30-patient test cohort. All rows report
measured wall time. The first two rows are the sequential pipeline under
single-patient and batched-across-patient regimes; the bottom row is the fully
agentic build (Section~2.1.4) on the held-out test30 cohort, including all chunk
extraction, unification, episode synthesis, and display generation stages.}
\label{tab:parallel}
\end{table*}

Graph operations via NetworkX executed in 0.1--1.5\,ms for temporal, episodic, and
measurement queries, with retrieval reduced to deterministic graph traversal over
the pre-computed TOA layer. End-to-end chat latency remained clinically responsive
at 1--3 seconds total (Stage 1 planning + Stage 2 graph execution $<$10\,ms +
Stage 3 narrative generation), with the bulk of wall time consumed by downstream
LLM generation rather than retrieval. The deterministic MEDS Graph construction
step itself completed in a median of 0.1\,s per patient (representative
large patient in Fig.~\ref{fig:build_performance}: $\sim$10{,}385 nodes
built in 0.6\,s with peak memory under 15\,MB; the cohort-wide maximum
record contained 136{,}508 nodes); every TOA
event linked to at least one source XML fragment, and provenance lookups
returned in under 0.01\,s. Coverage of six representative graph-resident retrieval
targets on the test30 cohort is reported in
Appendix~\ref{app:retrieval_coverage} (Supplementary Table~\ref{tab:retrieval_coverage}). For deployments larger than a single research
cohort, the same graph schema is compatible with Neo4j, enabling consolidation
into a population-scale graph with Cypher-based cohort queries.

\subsection{Graph-Guided Patient Retrieval}

A key advantage of structuring patient data into standardized, queryable artifacts is
the ability to perform cross-patient retrieval across the full cohort without
additional EHR reprocessing. As a demonstration, we built a \emph{patient similarity
module} that surfaces clinically relevant patients from the 1{,}180-patient cohort,
showing that a straightforward application of the pre-computed graph artifacts is
already a working retrieval system---not a final product, but an illustration that
the architecture readily supports cross-patient analyses without re-touching the raw
EHR. This capability operationalizes the emerging paradigm of \emph{smart patient
retrieval}---AI systems that surface similar prior cases with known outcomes to
provide contextual evidence on disease trajectories, treatment responses, and trial
opportunities during tumor board deliberation---offering a concrete starting point
for the kind of retrieval recently advocated for precision oncology.\cite{wang2026smart}

The module is not designed to maximize a single notion of overall similarity.
Different clinical questions privilege different facets of a patient's record: a
question about expected toxicity from a novel EGFR-MET combination requires patients
matched on driver biology; a question about whether to attempt salvage surgery after
progression on first-line immunotherapy requires patients matched on treatment
trajectory; a question about an unusual presentation pattern requires matching on
the narrative descriptions in the patient's pre-tumor-board summary note. The
module therefore exposes explicit weights over distinct retrieval components, so
that the same underlying graph can serve different retrieval objectives without
re-indexing.

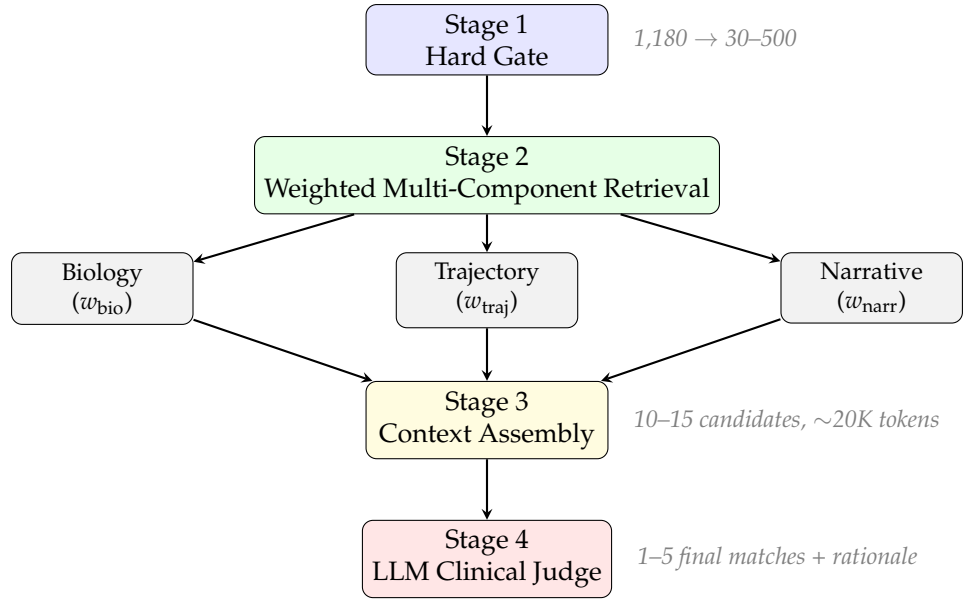
\begin{figure*}[!t]
\centering
\begin{tikzpicture}[
    node distance=0.6cm and 0.4cm,
    stage/.style={rectangle, draw, rounded corners, minimum width=3.2cm, minimum
height=0.9cm, align=center, font=\small},
    net/.style={rectangle, draw, rounded corners, minimum width=2.4cm, minimum
height=0.7cm, align=center, font=\footnotesize, fill=gray!10},
    arrow/.style={->, >=stealth, thick},
    label/.style={font=\footnotesize\itshape, text=gray}
]
    \node[stage, fill=blue!10] (gate) {Stage 1\\Hard Gate};
    \node[label, right=0.2cm of gate] {1{,}180 $\rightarrow$ 30--500};

    \node[stage, fill=green!10, below=0.8cm of gate] (retrieve) {Stage 2\\Weighted
Multi-Component Retrieval};

    \node[net, below left=0.5cm and 0.8cm of retrieve] (bio) {Biology\\($w_{\mathrm{bio}}$)};
    \node[net, below=0.5cm of retrieve] (traj) {Trajectory\\($w_{\mathrm{traj}}$)};
    \node[net, below right=0.5cm and 0.8cm of retrieve] (narr) {Narrative\\($w_{\mathrm{narr}}$)};

    \node[stage, fill=yellow!15, below=2.2cm of retrieve] (assemble) {Stage 3\\Context
 Assembly};
    \node[label, right=0.2cm of assemble] {10--15 candidates, $\sim$20K tokens};

    \node[stage, fill=red!10, below=0.8cm of assemble] (judge) {Stage 4\\LLM Clinical
Judge};
    \node[label, right=0.2cm of judge] {1--5 final matches + rationale};

    \draw[arrow] (gate) -- (retrieve);
    \draw[arrow] (retrieve) -- (bio);
    \draw[arrow] (retrieve) -- (traj);
    \draw[arrow] (retrieve) -- (narr);
    \draw[arrow] (bio) -- (assemble);
    \draw[arrow] (traj) -- (assemble);
    \draw[arrow] (narr) -- (assemble);
    \draw[arrow] (assemble) -- (judge);

\end{tikzpicture}
\caption{\textbf{Graph-guided patient similarity pipeline.} Stages~1--3 operate
entirely on pre-computed TOA artifacts without LLM calls. Three retrieval
components each score a candidate on a distinct facet of clinical similarity,
combined through tunable weights: $w_{\mathrm{bio}}$ weights biology (Jaccard
overlap of driver-mutation profiles, with histology and PD-L1 concordance);
$w_{\mathrm{traj}}$ weights treatment trajectory (agreement in line number,
therapeutic intent, and regimen drug-class); and $w_{\mathrm{narr}}$ weights
narrative (BM25 lexical match over each patient's pre-tumor-board summary note).
Because these weights are applied at query time over the same pre-computed index,
the graph can serve different retrieval objectives without re-indexing. Only
Stage~4 requires a single LLM invocation for final clinical judgment.}
\label{fig:plm_pipeline}
\end{figure*}

The pipeline operates in four stages (Figure~\ref{fig:plm_pipeline}).
\textbf{Stage~1 (Hard Gate)} filters the cohort by diagnosis category (NSCLC, SCLC,
mesothelioma, thymoma) and metastatic status, reducing 1{,}180 candidates to a
clinically compatible subset (typically 30--500 patients). \textbf{Stage~2 (Weighted
Multi-Component Retrieval)} applies three parallel scoring components, each
targeting a distinct facet of clinical similarity. The relative contribution of each
component is controlled by weights $w_{\mathrm{bio}}, w_{\mathrm{traj}},
w_{\mathrm{narr}}$, which for this demonstration were set to equal contribution
($w_{\mathrm{bio}} = w_{\mathrm{traj}} = w_{\mathrm{narr}} = 1/3$):

\begin{enumerate}[leftmargin=1.5em, itemsep=2pt]
  \item \textbf{Biology component:} Weighted Jaccard similarity on driver mutation
profiles (EGFR, ALK, KRAS, ROS1, BRAF, MET, RET, NTRK, HER2) with secondary terms
for histology and PD-L1 concordance. Matching actionable driver mutations (e.g.,
both EGFR-positive) receive 3$\times$ weight, reflecting their outsized influence on
treatment selection.
  \item \textbf{Trajectory component:} Composite score based on treatment line number,
therapeutic intent (curative vs.\ palliative), regimen drug class overlap (14 classes:
 platinum doublet, immunotherapy, EGFR TKI, etc.), and prior surgery/radiation status.
  \item \textbf{Narrative component:} BM25 ranking on each patient's pre-tumor-board
summary note. This note is generated by VISTA Architect at patient ingestion
following the standard Stanford pre-tumor-board summary procedure---a concise
clinical summary of the same format presented at current Stanford tumor board
meetings---and is stored as a graph-resident artifact alongside the structured
TOA layer. Because the note is part of the architecture and is generated for
every patient at ingestion time, it is available as a query input for any
patient (including the index patient, before that patient's actual board
meeting). The narrative component captures clinical descriptors not always
represented in structured fields (e.g., ``bilateral involvement,'' ``brain
metastases,'' ``poor functional status'').
\end{enumerate}

Each component returns its top-5 candidates; after deduplication, 10--15 candidates
advance to \textbf{Stage~3 (Context Assembly)}, where the index patient's full
clinical profile and each candidate's compressed profile are packed into a single
prompt ($\sim$15--20K tokens). \textbf{Stage~4 (LLM Clinical Judge)} uses a
reasoning model to evaluate candidates and select 1--5 final matches, providing
per-match rationale, key differences, and the matched patient's
post-tumor-board trajectory.

Critically, Stages~1--3 require \emph{no LLM calls}: the retrieval index is
pre-computed from the same \texttt{patient\_info.json}, \texttt{episodes.json}, and
\texttt{summary.json} artifacts generated during per-patient TOA construction. Index
construction for 1{,}180 patients completes in under 5~seconds, and retrieval
executes in under 100~ms per query. Only the final clinical judge step (Stage~4)
requires a single LLM call.

As an illustrative query, querying an EGFR-positive metastatic adenocarcinoma patient
on second-line palliative therapy retrieved 14 candidates from 554 after hard gating.
The top-ranked match---also EGFR-positive adenocarcinoma on second-line palliative
intent---achieved a biology score of 1.0 and trajectory score of 0.87, with
multi-component retrieval surfacing this patient through both the biology and
trajectory components independently.

The equal-weight setting used here is a default rather than a fixed choice. If a
specific retrieval task warranted emphasizing one facet over another---for example,
weighting biology more heavily when the clinical question concerns mutational
analogues, or weighting narrative when targeting an unusual presentation
pattern---the same underlying graph artifacts support arbitrary weight
configurations without re-indexing.

\section{Discussion}

VISTA Architect addresses critical challenges encountered by multidisciplinary tumor boards, particularly the difficulty of synthesizing fragmented clinical narratives derived from complex electronic health record (EHR) data.\cite{budd2023burnout,sinsky2016allocation,han2019estimating,schulte2019death} Traditional methods often require repetitive re-processing of extensive EHR records at query time, introducing substantial computational overhead and latency. In contrast, VISTA Architect structures clinical data into a persistent, temporally structured, and context-specific Timeline Object Architecture (TOA) data plane, inherently capturing critical information required for clinical decision-making. This approach reduces computational demands and enables near-instantaneous query response.

The architectural advantage of VISTA Architect is not that graph databases categorically replace relational storage. The lowest EHR layer could in principle be backed by SQL, document stores, or other clinical data systems. The key distinction is representational: VISTA transforms longitudinal documentation into a hierarchical, clinically navigable structure. The MEDS Graph preserves source-level detail; the TOA Graph abstracts that detail into temporally resolved clinical events; and the episode and current-state layers compress the event stream into clinically meaningful phases and decision-ready artifacts. Query-time reasoning therefore begins from an organized patient state and traverses downward only when source-level evidence is needed.

This changes the role of retrieval. A relational query can return rows satisfying predefined predicates, and a RAG system can retrieve text fragments matching a query. In VISTA, the TOA layer provides a clinical map of the record: diagnosis, treatment phases, progression events, imaging anchors, toxicities, and current state are represented as connected objects. The agent can therefore retrieve by following clinically meaningful edges and targeting relevant nodes, rather than scanning or reassembling the patient history from raw tables or documents. This resembles how clinicians navigate a chart: they begin with the known clinical story, then drill into the specific note, report, image, or lab value needed to verify a claim.

Viewed from an information-processing perspective, the architecture functions as a log-compressed clinical tree built to minimize information loss while maximizing navigability. The TOA and episode layers are structured compressions of the longitudinal EHR that preserve clinically salient state transitions, relationships, and provenance while organizing information at progressively higher levels of abstraction. For an average downstream query, many high-level facts are already present in context, and the model only needs to retrieve incremental details. This reduces the amount of text passed into the model, the amount of reasoning required to reconstruct patient state, and the number of source-level lookups, while retaining access to the original EHR when needed.

Because the representation is graph-based, new node types, edge types, provenance links, or specialty-specific episode definitions can be added without redefining the entire system. The same source-faithful graph can therefore support multiple clinical configurations and downstream tools, including dashboards, similarity search, automated note writing, plotting, and conversational agents.

Pre-computing the graph also exposes a clean substrate for agents. Because TOA
assertions are provenance-tracked and graph-resident, agents can operate over
deterministic, audited context without performing their own raw-text retrieval.
This allowed the build pipeline, dashboard generation, chat, and similarity
retrieval to share a single agentic interface (Section~2.1.4) while inheriting
the structural guarantees of the graph layer underneath it. This shared agentic interface was the configuration used to measure wall-time gains under full intra- and inter-patient parallelization without degradation of extraction quality (Section~2.5).

These performance characteristics should be contextualized against the broader landscape of LLM-based clinical information extraction. While LLMs have demonstrated strong capabilities for single-document summarization tasks where the input fits within the model's context window,\cite{van2024adapted} the challenge of extracting structured clinical variables from full longitudinal EHR records---spanning years of documentation and thousands of events---remains substantially harder. A recent meta-analysis of 56 studies evaluating LLM integration in oncology decision-making reported an average overall accuracy of 76.2\%, with diagnostic accuracy at 67.4\%.\cite{hao2025large} Agentic multi-agent approaches such as the Healthcare Agent Orchestrator achieved 84\% strict recall on high-importance tumor board facts across 71 patients,\cite{blondeel2025healthcare} and optimized clinical RAG pipelines for note-level extraction reported F1 scores of 0.79--0.90 depending on retrieval strategy.\cite{lopez2025clinical} VISTA Architect achieved 96.4\% accuracy across 17{,}700 evaluations in 1{,}180 patients on the longitudinal whole-record extraction task. In our matched comparison on a 30-patient subset, a standard BM25 RAG baseline using the same LLMs achieved approximately 67\% accuracy where VISTA Architect achieved 96.9\% (Section~2.3). These results indicate that the evaluation task is non-trivial---a standard RAG approach does not trivially solve it---and that VISTA Architect's high accuracy is not an artifact of an easy benchmark.

VISTA Architect's demonstrated thoracic tumor board use case can be viewed as an automation and extension of a summarization workflow already developed and deployed at our institution:\cite{elliscaleo2026mtb} whereas that system summarizes a fixed lookback window of recent notes into a single live artifact, the precomputed, provenance-linked timeline yields the same summary note as one of several reusable artifacts, alongside structured variables and cross-patient retrieval. That deployment experience also indicates that, for well-specified summarization tasks, tightly constrained workflows can match or exceed higher-autonomy agentic designs; VISTA Architect is consistent with this principle, as its agentic layer operates over a deterministic, provenance-tracked graph rather than performing unconstrained retrieval, and its fully agentic build is used to parallelize processing without altering extraction accuracy (Section~2.3).

While VISTA Architect was conceived with general clinical workflows in mind, the current implementation and validation have been demonstrated within thoracic oncology tumor boards. The architecture supports flexibility in defining new node and edge types through configurable profiles; however, further empirical validation in additional clinical contexts remains necessary to substantiate broader claims about specialty agnosticism.

The present evaluation is based on a single-center implementation using exclusively English-language documentation. Future expansions to multiple centers, varied healthcare settings, and multilingual clinical documentation will be important to comprehensively assess generalizability and robustness. Additionally, the architecture's reliance on accurate and complete original EHR data is a notable limitation, as incomplete documentation inherently impacts the accuracy and completeness of the derived clinical narratives. A related limitation is that OMOP-derived data exports do not capture all sources that clinicians reference in practice---Care Everywhere notes, external lab results, and outside-institution records may be absent. This limitation is partially mitigated by our evaluation cutoff: ground truth was truncated at tumor-board-date minus one day, which guarantees that the available record was at minimum sufficient for the actual tumor board meeting to have proceeded. Downstream events occurring outside the index institution (e.g., mortality ascertainment or recurrence detection) are more substantially affected by these coverage gaps and would require complementary data sources.

Future directions for VISTA Architect include expanding validation across multiple sites, integrating multimodal data such as imaging and genomic information, and enhancing capabilities for managing larger patient cohorts. Prospective studies evaluating impacts on clinical workflow efficiency, user satisfaction, and clinical decision quality are also planned to provide a comprehensive assessment of the system's clinical utility. Several extensions are deferred to future work: ablation studies isolating the contribution of each TOA layer; explicit validation of the temporal-normalization step against chart-reviewed event dates; expanded clinician adjudication beyond the 30-patient calibration subset; and stronger retrieval baselines including dense-embedding and long-context single-pass configurations. The BM25 comparison reported in Section~\ref{subsec:rag_comparison} is best interpreted as a coarse graph-versus-no-graph ablation rather than as a head-to-head against an optimized clinical RAG system.

In summary, VISTA Architect demonstrates a pre-computation paradigm for integrating large, complex document repositories with LLM-driven applications: by constructing a structured, temporally aware knowledge graph once from raw source data, the system transforms the retrieval problem from a per-query search into a deterministic graph traversal, addressing both the accuracy and the speed requirements of high-stakes decision-making.\cite{hammer2020digital,nobori2022electronic,chang2025use} While validated here in thoracic oncology tumor boards, the underlying architecture---configurable node types, episode structures, and provenance-tracked event timelines---may be applicable to other settings where longitudinal, heterogeneous document collections must be made reliably accessible to AI systems; such applications will require prospective validation in their own clinical contexts.

\section{Methods (Evaluation)}

\subsection{Dataset}

Our evaluation cohort comprised 1{,}180 thoracic oncology patients drawn from the
\emph{VISTA Oncology Data Lake}, a multi-modal clinical data resource established
by Stanford Medicine's Research Technology group (full description in
Appendix~\ref{app:data_lake_note}\footnote{Public overview:
\url{https://susom.github.io/starr-oncology-data-lake-arpah/about.html}.}). The
Data Lake integrates clinical notes, genomic data, and imaging metadata
together with the Stanford Cancer Registry, covering approximately 222{,}000
patients across the Stanford hospital ecosystems; the derivation of the
1{,}180-patient evaluation cohort is summarized in Supplementary
Fig.~\ref{fig:cohort_flow}. Patient inclusion in the Data
Lake requires either a documented tumor board encounter in the source EHR or a
record in the Stanford Cancer Registry (which feeds the California Cancer
Registry). The clinical dataset is delivered as BigQuery tables transformed from
the source Epic Clarity instances to the Observational Medical Outcomes
Partnership (OMOP) Common Data Model, derived from Stanford's STARR-OMOP
electronic health record representation. For VISTA Architect, each patient's
OMOP-formatted record was exported from the Data Lake and converted into the
MEDS XML representation (Section~2.1) that serves as input to the MEDS Graph.
All records are Safe Harbor de-identified
by Research Technology before extraction; in particular, all dates (including
dates of birth, death, and service dates) are shifted by a per-patient offset
of $\pm$30 days (excluding zero), which preserves intra-patient temporal
ordering while removing the original calendar dates. No protected health
information left the institutional perimeter during analysis (Stanford IRB
protocol \texttt{76049}). All LLM inference was carried out
through PHI-compliant services---Stanford SecureGPT (Azure OpenAI) for the
sequential pipeline and Google Vertex AI for the agentic pipeline---so that
patient data never left the institutional data perimeter at any stage.

\subsection{Accuracy Assessment}

We evaluated VISTA Architect's clinical accuracy using a structured, variable-level
evaluation protocol across all 1{,}180 patients. From each patient, we extracted 15
MTB-salient variables spanning six clinical categories: demographics (Date of Birth,
Sex, Smoking Status), tumor characteristics (Diagnosis, Histology, Metastasis, Lymph
Node Involvement, Genetic Testing Panel), clinical status (ECOG Performance Status,
Therapy Toxicity/Comorbidities), treatment (Previous Surgery, Current Medical Therapy,
 Radiation Therapy), imaging (Date of Last CT), and safety (Allergies). This
yielded 17{,}700 total variable evaluations.

The variable set was designed to span both the structured--unstructured spectrum
and the range of extraction difficulty encountered in tumor board preparation,
from variables resolvable from a single field to variables that require careful
collection across the entire EHR history. Although several variables (notably
Date of Birth and Sex) are reliably available in structured OMOP fields and
could in principle be extracted deterministically, we deliberately designed the
accuracy assessment so that \emph{every} reported value was produced by the
generative pipeline: this exposes any hallucination upstream of the
note-extraction challenge, with the structured variables serving as positive
controls (a value other than 100\% accuracy on Date of Birth or Sex would
indicate a regression in the pipeline). The remaining variables fall into two
qualitatively different difficulty regimes. \emph{Collection variables}---Previous
Surgery, Therapy Toxicity / Comorbidities, and Allergies---require sweeping the
full longitudinal record and assembling a set; for Previous Surgery in
particular, the evaluation deliberately includes non-oncological and remote
surgical history (e.g., prior pediatric procedures relevant to anesthesia or
current surgical planning), which typically appears only in free-text past
medical history rather than in structured procedure tables. \emph{Time-varying
state variables}---ECOG Performance Status, Current Medical Therapy, Metastasis,
Lymph Node Involvement, and Date of Last CT---require extracting the current
status from multiple potentially conflicting mentions, where the temporal
development of the variable matters. The 15 variables were selected to have a
defined value for every oncology patient; absence of documentation is itself an
informative value (e.g., no documented oncologic surgery is interpreted as ``No
Previous Surgery''), and the LLM-as-judge scored ``no documentation''
symmetrically against the ground-truth XML.

To establish ground truth, we used each patient's raw EHR XML truncated at the
documented tumor board date, ensuring evaluation reflects only information available
at the clinical decision point. We employed an LLM-as-judge approach\cite{zheng2023llmjudge} using GPT-5 to
perform automated, bidirectional comparison between VISTA Architect's structured JSON
outputs and the XML ground truth. The judge assigned each variable a correctness label
 (Correct or Incorrect) and a 10-point quality score with detailed rationale. This
approach has been validated in clinical settings, where LLM-based fact verification
against EHR data achieves agreement with clinicians exceeding inter-clinician
agreement.\cite{chung2025verifact} An LLM-as-judge fact-scoring pipeline using the
same model family (GPT-5) was independently validated against physician fact-scoring
for thoracic tumor board summaries at the same institution, reporting judge--physician
agreement comparable to inter-physician agreement,\cite{elliscaleo2026mtb} supporting
its use as a scalable evaluator in this setting. To calibrate the LLM-as-judge in our setting, a
clinician independently reviewed all 480 variable evaluations across the randomly
selected, representative 30-patient subset described above, classifying each judge
decision as Agree, Uncertain, or Disagree against chart-reviewed ground truth. All
30 patients were evaluated using the same 16-variable schema; variables without
explicit documentation were retained as valid evaluable states when clinically
appropriate, such as the absence of documented oncologic surgery being interpreted
as no Previous Surgery rather than as missing data. The evaluation pipeline uses
distinct models across stages to avoid same-model self-evaluation: GPT-4.1 is used
for chunk-level TOA event extraction (where parallelization speed is prioritized),
while GPT-5 is used for downstream patient-info synthesis and, independently, as
the LLM judge for the accuracy evaluation reported here.

The same randomly selected, representative 30-patient subset was additionally used as the held-out test cohort for benchmarking the fully agentic parallelized build (Section~2.1.4) and the RAG retrieval baseline (Section~\ref{subsec:rag_comparison}, Appendix~\ref{app:rag_baseline}); no development of either system was conducted on this subset.

Our scoring rubric balanced strictness with clinical realism. A score of 10 required
exact or clinically equivalent matches between extracted and ground-truth values.
Partial credit (scores 5--9) was assigned for clinically accurate but incomplete
information---for example, correctly identifying adenocarcinoma histology but missing
a specific mutation subtype. We adopted specialized handling for ambiguous clinical
scenarios: metastasis and lymph node involvement were scored as Correct when imaging
suggested disease without pathological confirmation, reflecting real-world MTB
practice where treatment decisions must proceed on the basis of radiologic findings.

The large evaluation cohort (N=1{,}180) provides statistical power to estimate
accuracy with narrow 95\% confidence intervals. For mean scores, we report
normal-approximation CIs; for proportion correct, we report Wilson score intervals.
The scale of evaluation also mitigates a key concern with LLM-as-judge paradigms: at
small sample sizes, correlated hallucinations between extraction and evaluation models
 could systematically inflate accuracy. At N=1{,}180, such random concordance errors
contribute negligible bias ($<$0.1 percentage points), and systematic aligned errors
are further mitigated by the architectural separation between extraction
(graph-serialized input, deterministic retrieval context) and evaluation (raw XML
ground truth, independent judge prompt).

\paragraph{Decoding and determinism.} All LLM calls in the production pipeline
(GPT-4.1 chunk extraction, GPT-5 patient-info synthesis, GPT-5 LLM-as-judge) and
in the agentic pipeline (Gemini 3.5 Flash per-chunk extraction and episode
synthesis, Claude Opus 4.6 orchestrator / display) use vendor-default decoding
settings: temperature 1.0, top-$p$ 1.0, no explicit random seed, and provider
default \texttt{max\_tokens}. Reasoning-model parameters (\texttt{reasoning\_effort}
for GPT-5, \texttt{thinking\_budget} for Gemini 3.5 Flash) are not adjusted from
their defaults except that \texttt{thinking\_budget} is explicitly set to 0 for
short-output Flash calls to avoid token starvation. No determinism guarantees
are made or required for the reported accuracy numbers, which are reported as
single-run point estimates with binomial / Wilson confidence intervals
appropriate to the evaluation sample size (Table~\ref{tab:model_provenance}).

A completed TRIPOD-LLM reporting checklist (Gallifant et~al., 2025) is provided
as Supplementary File~S1, mapping each checklist item to the manuscript section
in which it is addressed.

\begin{table*}[t]
\centering
\small
\begin{tabular}{@{}p{3.6cm}p{2.8cm}p{6.2cm}p{2.6cm}@{}}
\toprule
\textbf{Pipeline role} & \textbf{Model identifier} & \textbf{Serving endpoint} &
\textbf{Access window} \\
\midrule
Chunk-level event extraction (sequential pipeline) &
  \texttt{gpt-4.1} & Stanford SecureGPT (Azure OpenAI; PHI-compliant) &
  Feb 2026 \\
Patient-info synthesis (sequential pipeline) &
  \texttt{gpt-5} & Stanford SecureGPT (Azure OpenAI) & Feb 2026 \\
LLM-as-Judge (accuracy evaluation) &
  \texttt{gpt-5} & Stanford SecureGPT (Azure OpenAI) & Feb 2026 \\
RAG baseline answer generation (variant 1) &
  \texttt{gpt-4.1} & Stanford SecureGPT (Azure OpenAI) & Apr 2026 \\
RAG baseline answer generation (variant 2) &
  \texttt{gpt-5} & Stanford SecureGPT (Azure OpenAI) & Apr 2026 \\
Per-chunk event extraction \& episode synthesis (agentic pipeline) &
  \texttt{gemini-3.5-flash} & Google Vertex AI (Stanford GCP project
  \texttt{som-nero-plevriti-deidbdf}) & May--Jun 2026 \\
Agentic orchestrator \& display synthesis (agentic pipeline) &
  \texttt{claude-opus-4-6} & Google Vertex AI (Stanford GCP project
  \texttt{som-nero-plevriti-deidbdf}) & May--Jun 2026 \\
\bottomrule
\end{tabular}
\caption{Exact model identifiers, serving endpoints, and access windows for
every LLM used in this study. Both endpoints are PHI-compliant: SecureGPT
proxies Azure OpenAI inside the Stanford institutional perimeter; the
Vertex~AI project is a Stanford-managed GCP environment under the same
data-use agreement. Decoding settings for all models are vendor defaults
(\S4.2 ``Decoding and determinism''). Model identifiers are pulled verbatim
from \texttt{gsgpt.py} (\texttt{MODELS}, \texttt{VERTEX\_MODELS},
\texttt{VERTEX\_ANTHROPIC\_MODELS}) and from the \texttt{gcpclaude} launcher
environment.}
\label{tab:model_provenance}
\end{table*}

\subsection{Efficiency Benchmarks}

We measured VISTA Architect's computational performance at both the cohort and
single-query level. At the cohort level, we measured end-to-end build wall time
under two configurations: the sequential pipeline used to generate all results in
Section~2.3 (running on one patient at a time, and additionally with 10 patients
in parallel as the production regime), and the fully agentic pipeline
(Section~2.1.4) running with full intra- and inter-patient parallelization. At the
single-query level, we measured graph query latency across temporal, episodic,
and measurement operations on the pre-computed TOA layer, and end-to-end chat
latency encompassing all three stages of the agentic chat pipeline.

\section*{Code and Data Availability}

The VISTA Architect backend---comprising MEDS XML to MEDS Graph conversion, TOA
event extraction, episode synthesis, and graph-resident query infrastructure---is
available as open source on GitHub at
\url{https://github.com/VISTA-Stanford/vista-architect}, together with example
prompts sufficient to apply the pipeline to MEDS-formatted data from any source; the
snapshot corresponding to this manuscript is tagged \texttt{v0.1.0-preprint} (commit
\texttt{8837d44}). The agentic implementation of the AI layer (Section~2.1.4,
Appendix~\ref{app:agentic_implementation}) is released alongside the backend in the
same repository. The production frontend (Dash/Plotly user interface) is not part of
this release; the exact production prompts used for the thoracic oncology
configuration are reproduced verbatim in Supplementary File~S2. The
\texttt{meds2text} library used in Step~1 is publicly available at
\url{https://github.com/VISTA-Stanford/meds2text}.

The VISTA Oncology Data Lake from which the cohort was drawn is governed by
Stanford Medicine TDS under the ARPA-H program; access procedures and eligibility
are described at
\url{https://susom.github.io/starr-oncology-data-lake-arpah/about.html}.
Patient-level data cannot be redistributed by the authors.

\section*{Acknowledgments}
We thank the Stanford thoracic oncology team for clinical guidance throughout the
development of VISTA Architect. We acknowledge Stanford Medicine's Technology and
Digital Solutions (TDS) group, and in particular Somalee Datta, for building and
maintaining the VISTA Oncology Data Lake from which the cohort used in this study
was drawn. This research was funded, in part, by the Advanced Research Projects Agency for
Health (ARPA-H). The views and conclusions contained in this document are those
of the authors and should not be interpreted as representing the official
policies, either expressed or implied, of the U.S. Government. T.K. was supported by the
Finnish Cultural Foundation.

\section*{Competing Interests}
The authors declare no competing financial or non-financial interests.

\section*{Author Contributions (CRediT)}
Conceptualization: T.K., M.A.R., S.P., J.F., P.A., D.W., T.J.E.-C., A.F., B.N., and J.N.
Methodology: T.K., J.F., D.W., T.J.E.-C., and J.N. Software: T.K. and J.F.
Validation: T.K., A.F., D.W., T.J.E.-C., and J.N. Formal analysis: T.K. Investigation: T.K. and P.A.
Resources: J.F., P.A., B.N., S.P., and M.A.R. Data curation: T.K., J.F., P.A., and
B.N. Visualization: T.K. Supervision: M.A.R. and S.P. Project administration:
B.N., S.P., and M.A.R. Funding acquisition: S.P. and M.A.R. Writing -- original
draft: T.K. and M.A.R. Writing -- review and editing: all authors.

\section*{Ethics Approval}
This study was reviewed and approved by the Stanford University Institutional
Review Board under protocol \texttt{76049} (same value as
\S4.1). The protocol covers secondary research use of de-identified EHR data
extracted from the VISTA Oncology Data Lake. No protected health information
left the institutional perimeter at any stage of analysis. The Data Lake is
Safe Harbor de-identified by Stanford Medicine Research Technology before
extraction, with per-patient date shifts of $\pm$30 days (excluding zero) to
preserve intra-patient temporal ordering while removing the original calendar
dates. Patient consent was waived in accordance with 45~CFR~46.116(f) on the
basis of minimal risk and infeasibility of obtaining individual consent for
secondary use of de-identified records.

\bibliographystyle{unsrt}
\bibliography{references}

\clearpage
\newpage

\clearpage
\nolinenumbers
\onecolumn
\appendix

\renewcommand{\thesection}{\Alph{section}}
\setcounter{figure}{0}
\renewcommand{\thefigure}{S\arabic{figure}}
\setcounter{table}{0}
\renewcommand{\thetable}{S\arabic{table}}

\section*{Supplementary Information}
\addcontentsline{toc}{section}{Supplementary Information}

\bigskip
{\large\sffamily\bfseries Supplementary Contents}
\begin{itemize}[label={},leftmargin=0pt,itemsep=3pt]
\item \textbf{\ref{app:supp_figures}.}~~\nameref{app:supp_figures}\dotfill\pageref{app:supp_figures}
\item \textbf{\ref{app:data_lake_note}.}~~\nameref{app:data_lake_note}\dotfill\pageref{app:data_lake_note}
\item \textbf{\ref{app:meds_schema}.}~~\nameref{app:meds_schema}\dotfill\pageref{app:meds_schema}
\item \textbf{\ref{app:toa_schema}.}~~\nameref{app:toa_schema}\dotfill\pageref{app:toa_schema}
\item \textbf{\ref{app:prompts}.}~~\nameref{app:prompts}\dotfill\pageref{app:prompts}
\item \textbf{\ref{app:variables}.}~~\nameref{app:variables}\dotfill\pageref{app:variables}
\item \textbf{\ref{app:rag_baseline}.}~~\nameref{app:rag_baseline}\dotfill\pageref{app:rag_baseline}
\item \textbf{\ref{app:agentic_implementation}.}~~\nameref{app:agentic_implementation}\dotfill\pageref{app:agentic_implementation}
\end{itemize}
\clearpage
\newpage

\section{Supplementary Figures}
\label{app:supp_figures}

\begin{figure}[H]
\centering
\begin{tikzpicture}[
    node distance=10mm,
    box/.style={rectangle, draw, rounded corners, minimum width=8.6cm,
                minimum height=1.1cm, align=center, font=\small, inner sep=5pt},
    arrow/.style={->, >=stealth, thick}
]
    \node[box, fill=blue!8] (a) {\textbf{VISTA Oncology Data Lake}\\$\sim$222{,}000 patients (Stanford Medicine, all malignancies)};
    \node[box, fill=blue!14, below=of a] (b) {\textbf{Thoracic tumor board cohort}\\documented thoracic tumor board encounter; first\\encounter on/after 1 January 2020 (de-identified)\\$n=1{,}180$ \,(15 variables; 17{,}700 evaluations)};
    \node[box, fill=green!16, below=of b] (c) {\textbf{Validation subset}, $n=30$\\clinician concordance review, agentic build\\benchmark, and BM25 RAG comparison};

    \draw[arrow] (a) -- (b);
    \draw[arrow] (b) -- (c);
\end{tikzpicture}
\caption{\textbf{Cohort flow.} Selection of the 1{,}180-patient thoracic oncology
evaluation cohort from the VISTA Oncology Data Lake, and of the 30-patient subset
used for clinician concordance review, the agentic build benchmark, and the BM25
RAG comparison. The full cohort contributes 15 tumor board--salient variables per
patient (17{,}700 evaluations).}
\label{fig:cohort_flow}
\end{figure}
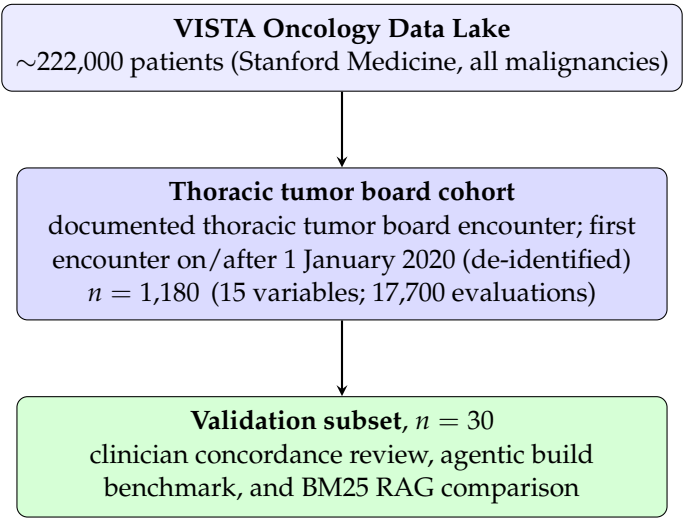

\begin{figure}[H]
  \centering
  \includegraphics[width=\textwidth]{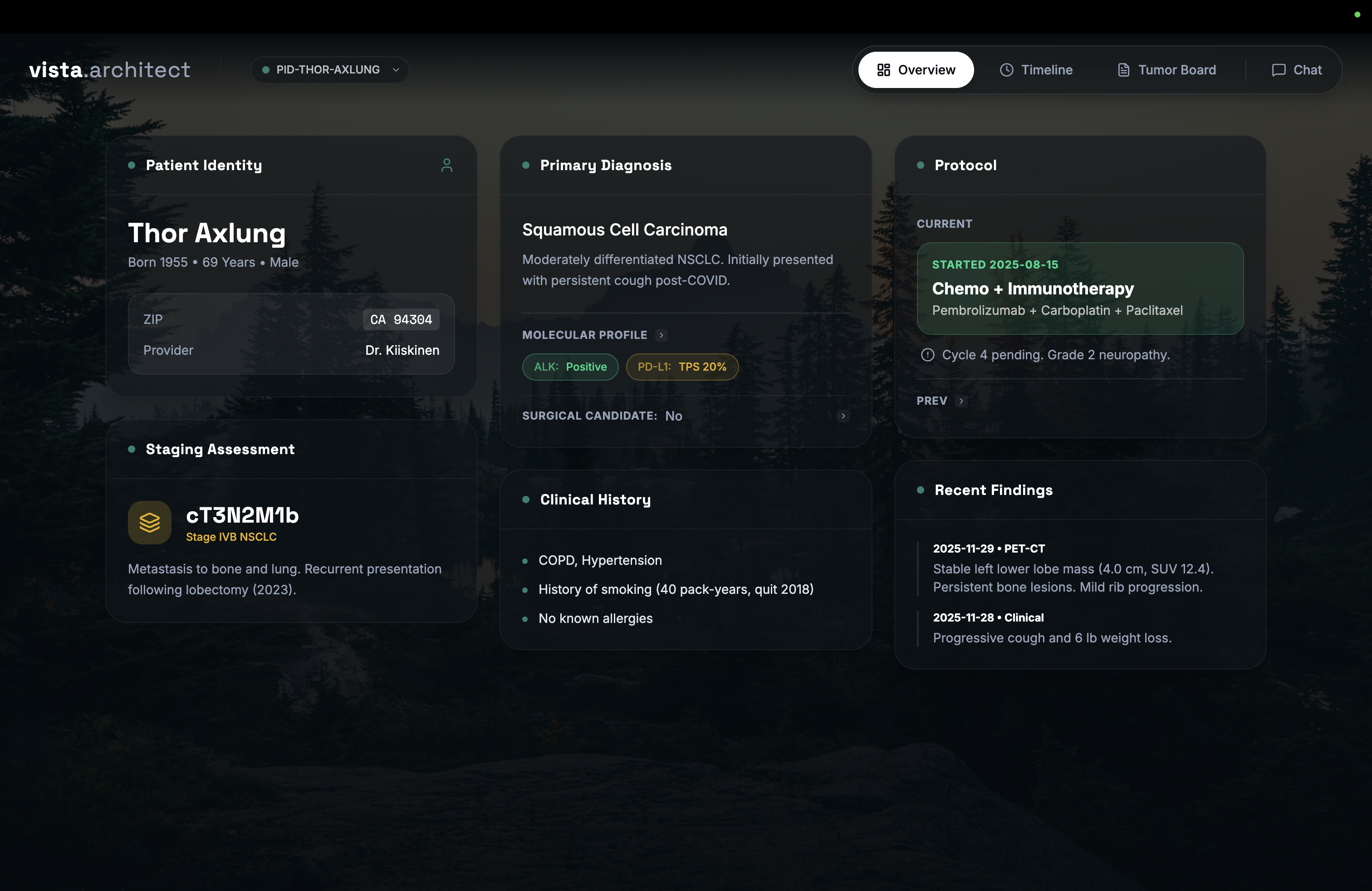}
  \caption{\textbf{VISTA Architect Overview view.} A single-screen synopsis of the
  patient assembled from graph-resident artifacts: patient identity, staging
  assessment, primary diagnosis with molecular profile, current and prior treatment
  protocol, clinical history, and the most recent findings. All fields are populated
  from the TOA layer and the structured \texttt{patient\_info.json} and
  \texttt{summary.json} artifacts produced during pipeline processing. Synthetic
  demonstration patient; not a real patient.}
  \label{fig:ui_overview}
\end{figure}

\begin{figure}[H]
  \centering
  \includegraphics[width=\textwidth]{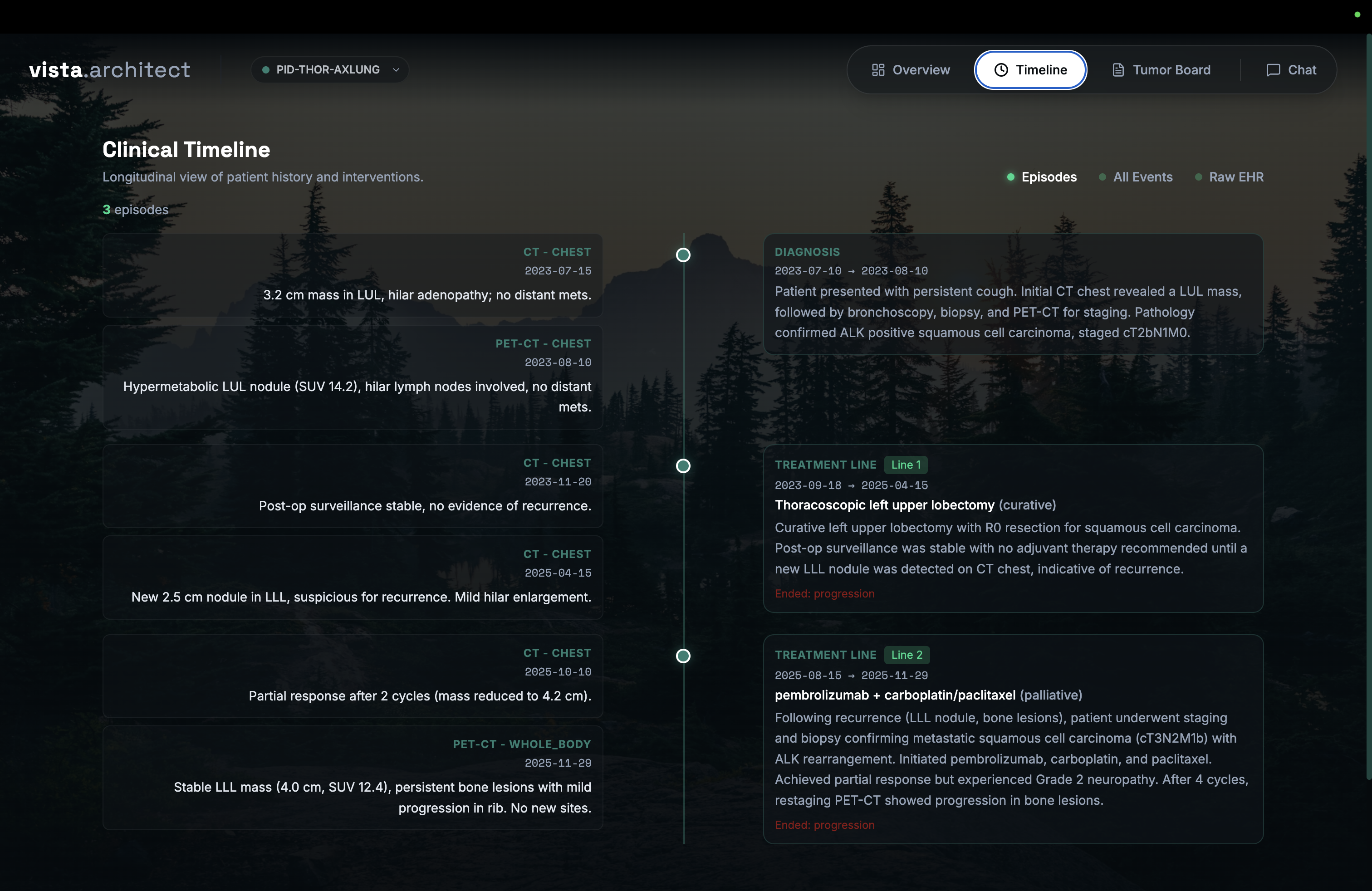}
  \caption{\textbf{VISTA Architect Clinical Timeline view.} The longitudinal record
  organized into clinical episodes---diagnosis and numbered treatment lines, each with
  start/end dates, intent, and termination reason---on the right, with the contributing
  imaging and assessment events aligned chronologically on the left. Toggles switch
  between the episode-level summary, the complete event log, and the raw EHR. The view
  is rendered directly from TOA episodes and events, preserving temporal structure and
  source provenance. Synthetic demonstration patient; not a real patient.}
  \label{fig:ui_timeline}
\end{figure}

\begin{figure}[H]
  \centering
  \includegraphics[width=\textwidth]{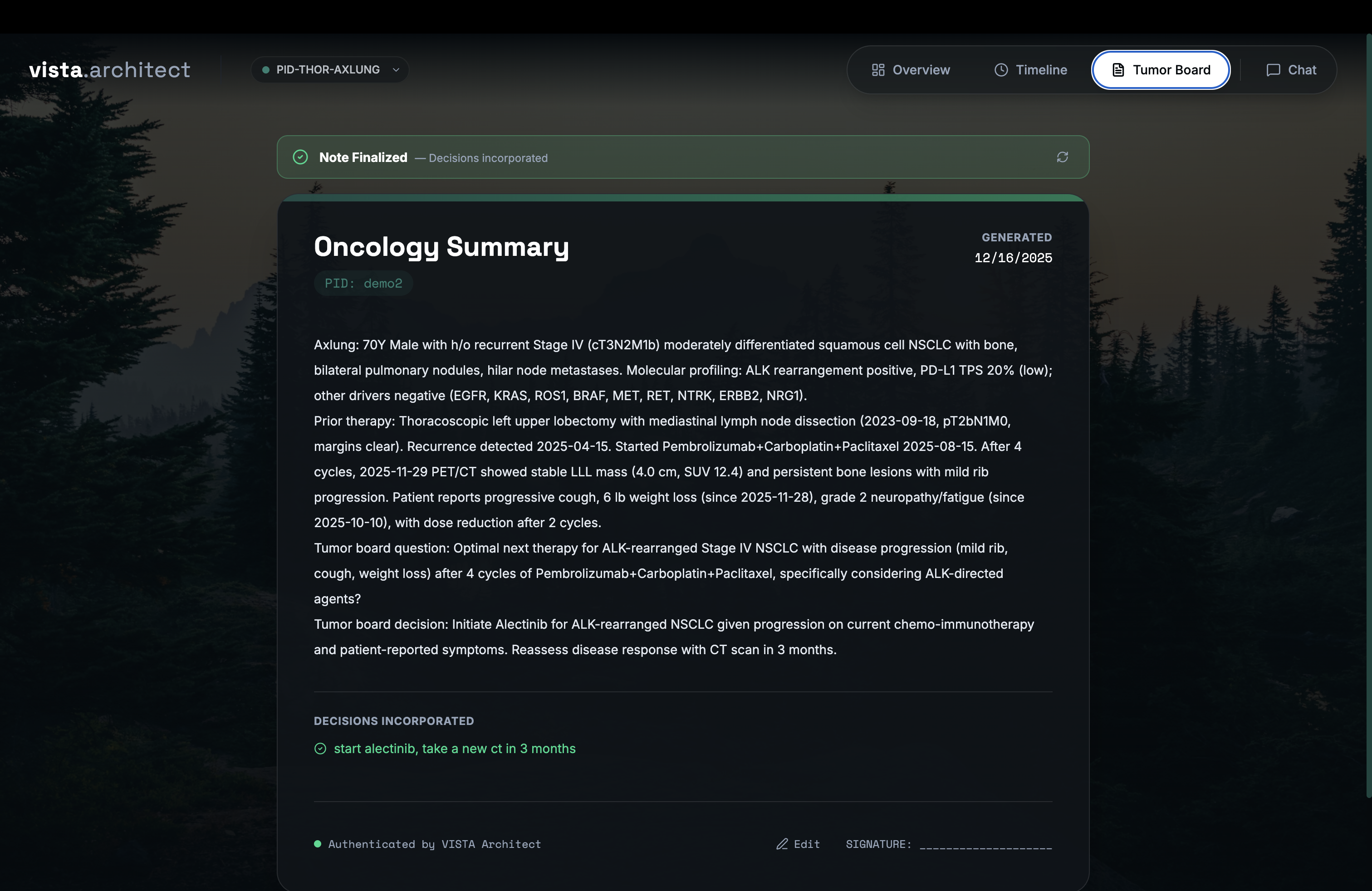}
  \caption{\textbf{VISTA Architect automated tumor board note.} Automated
  pre-tumor-board summary note writing incorporates the whole patient history
  together with recorded tumor board decisions, generated from the same
  graph-resident artifacts. Synthetic demonstration patient; not a real patient.}
  \label{fig:ui_note}
\end{figure}

\begin{figure}[H]
  \centering
  \includegraphics[width=\textwidth]{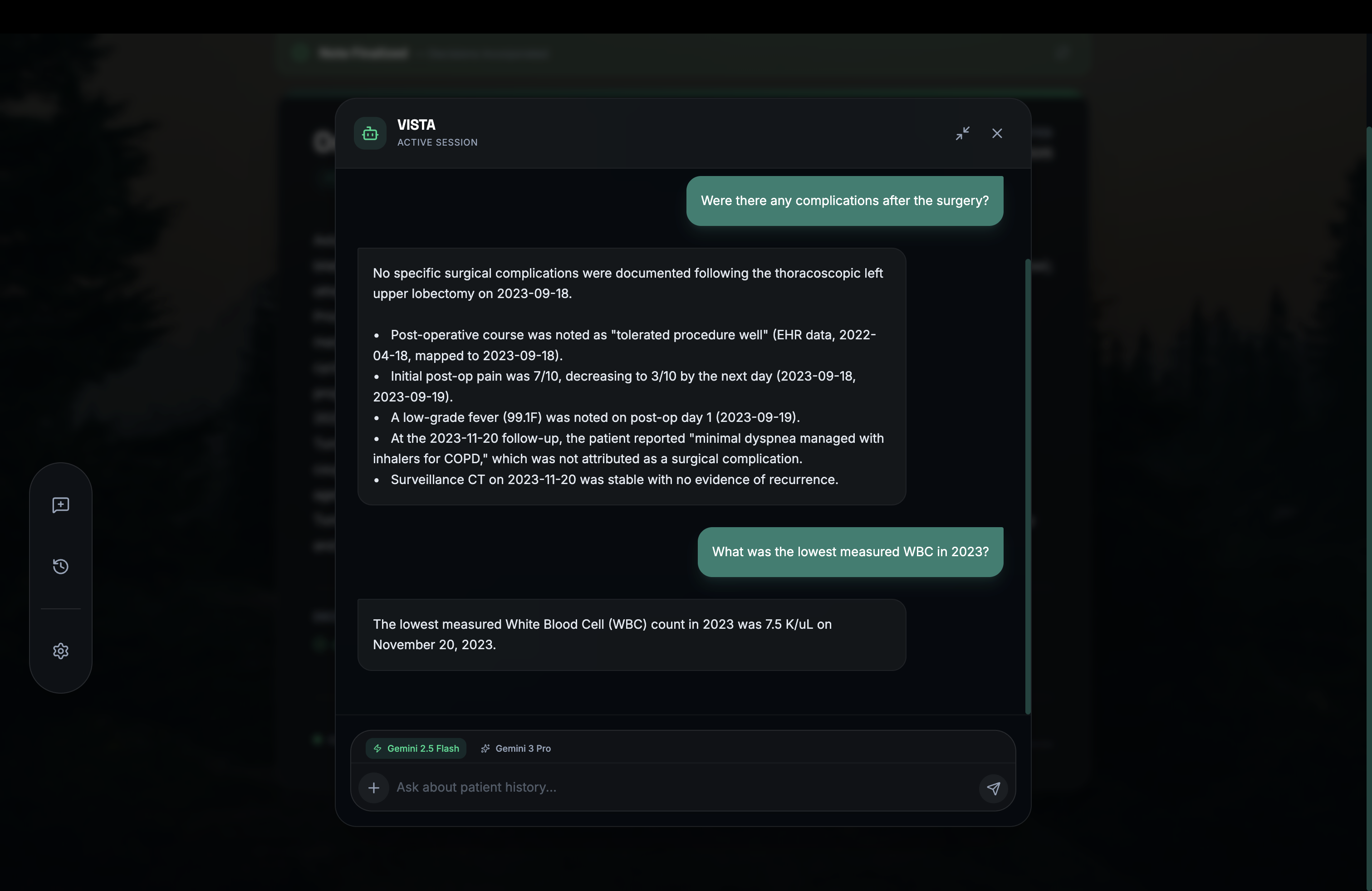}
  \caption{\textbf{VISTA Architect chat interface.} The chat interface answers queries
  either directly from the TOA data plane or by reaching into the MEDS Graph data
  plane through targeted agentic graph queries, with provenance-linked access to the
  source EHR. Synthetic demonstration patient; not a real patient.}
  \label{fig:ui_chat_interface}
\end{figure}

\begin{figure}[H]
  \centering
  \includegraphics[width=\textwidth]{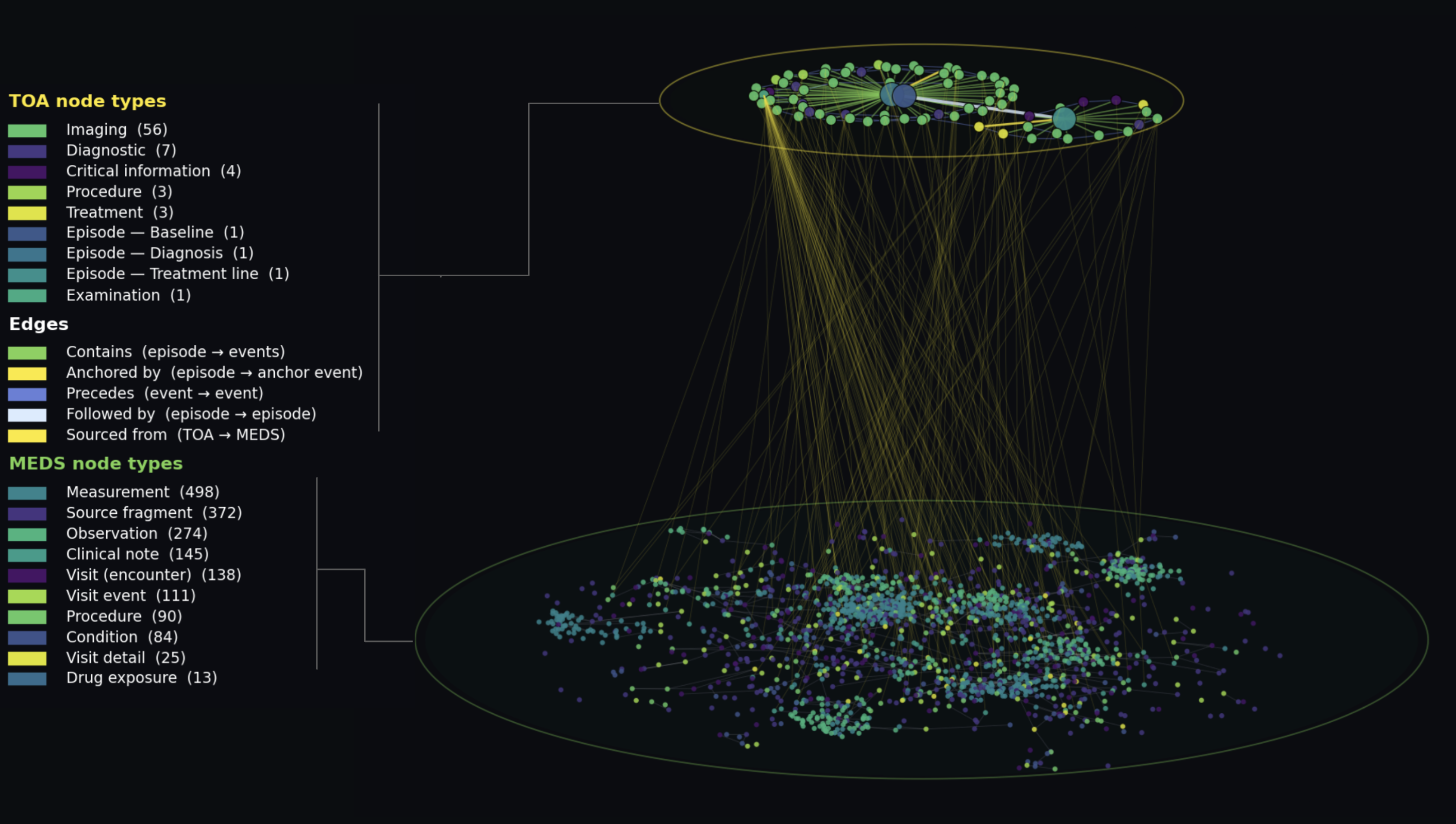}
  \caption{\textbf{Two-tier graph structure for a single representative patient.}
  Lower plane: the source-faithful MEDS Graph, whose nodes preserve the original EHR
  records (here dominated by measurements, source fragments, observations, clinical
  notes, and visits). Upper plane: the abstracted TOA Graph, in which clinical events
  are grouped into a small number of episode nodes (baseline, diagnosis, treatment
  line). Gold \emph{sourced-from} edges link every TOA event back to its supporting
  MEDS evidence, making provenance explicit; \emph{contains}, \emph{anchored-by},
  \emph{precedes}, and \emph{followed-by} edges organize the clinical timeline. Legend
  labels denote the \emph{semantic meaning} of each node and edge in this
  configuration, not a fixed type vocabulary (Section~2.1.3). For this patient the
  $\sim$1{,}750 MEDS-graph nodes are abstracted into $\sim$77 TOA nodes; counts are
  illustrative of a single smaller-than-median patient (cohort median 3{,}608
  MEDS-graph nodes).}
  \label{fig:graph_structure}
\end{figure}

This section collects the supplementary figures referenced throughout the main
text. Supplementary Fig.~\ref{fig:cohort_flow} summarizes selection of the
evaluation cohort. Supplementary Figs.~\ref{fig:ui_overview}, \ref{fig:ui_timeline},
\ref{fig:ui_note}, and~\ref{fig:ui_chat_interface} illustrate the VISTA Architect
user interface---the Overview, Clinical Timeline, automated pre-tumor-board summary
note, and chat views---each rendered directly from graph-resident artifacts with
provenance-linked access to the source EHR. Supplementary
Fig.~\ref{fig:graph_structure} shows the two-tier MEDS\,+\,TOA graph structure for a
single representative patient.

\clearpage
\section{VISTA Oncology Data Lake: Supplementary Note}
\label{app:data_lake_note}

This appendix provides additional context on the VISTA Oncology Data Lake, the
clinical data resource from which the 1{,}180-patient evaluation cohort in this
study was drawn. The condensed version of this material in Section~4.1 covers
the points directly relevant to the analyses reported in this paper; the
description here documents the broader scope, source systems, and operational
properties of the resource for readers seeking institutional context.

\paragraph{Scope and modalities.} The Virtual Intelligence for Specialized Tumor
Board Assessment (VISTA) Oncology Data Lake is a multi-modal data repository
designed to integrate diverse clinical modalities---including clinical notes,
genomic data, and imaging data---to facilitate oncology research and AI
modeling. To bridge fragmentation of care across different clinics, the Data
Lake also integrates the Stanford Cancer Registry. The Data Lake contains
approximately 222{,}000 patients in total; approximately 12\% of patients in
the Stanford Cancer Registry have a documented tumor board encounter.

\paragraph{Cohort inclusion criteria.} A patient is included in the Data Lake if
at least one of the following is true: (i) a documented tumor board presentation
or evaluation in the source Electronic Health Record (EHR); or (ii) a verified
analytic or non-analytic case in the Stanford Cancer Registry (SCR). The SCR
tracks comprehensive disease histories dating back to 1988 and feeds into the
California Cancer Registry. SCR analytic cases include patients whose first
course of cancer treatment occurred at Stanford regardless of diagnosis
location, and patients diagnosed at Stanford regardless of where they were
subsequently treated (with potentially incomplete data in the latter case).
SCR non-analytic cases---introduced in 2024---capture additional cancer-related
encounters such as patients treated at Stanford for recurrence only, patients
followed at Stanford and found to be cancer-free, and consultation-only or
end-of-life-care patients neither diagnosed nor treated at Stanford.

\paragraph{Source systems.} Source EHR data comes from the three Stanford
hospital ecosystems---Stanford Health Care, Stanford Children's Hospital
(formerly Lucile Packard Children's Hospital), and Stanford Healthcare
Tri-Valley---together with the network of more than one hundred pediatric and
adult care clinics affiliated with University HealthCare Alliance and Packard
Children's Health Alliance. These ecosystems use two independent Epic instances
that share Epic Clarity data with Research Technology. Non-Epic clinical sources
(radiology DICOMs, whole-slide imaging, genetic-testing results) are shared
across the two hospital ecosystems and are processed from single source
systems. The Stanford Cancer Registry is shared across both ecosystems and
managed on the Neuralframe KACI vendor platform, with monthly database
snapshots delivered to Research Technology.

\paragraph{Data architecture.} The Data Lake is delivered as BigQuery datasets
on Google Cloud Platform. BigQuery is a fully managed, serverless,
HIPAA-compliant enterprise data platform optimized for high-performance
analytics on large multi-modal biomedical datasets, with decoupled storage and
compute enabling complex SQL queries in seconds. Two independent BigQuery
datasets (EHR and Registry) are linked by the unique patient identifier
(Medical Record Number and Date of Birth). The clinical dataset is transformed
from the two Epic Clarity instances to the unified Observational Medical
Outcomes Partnership (OMOP) Common Data Model; the data includes Epic
modules such as Epic Genetics, Epic Beaker, and Epic Beacon. DICOM and WSI
metadata are included in custom tables alongside the OMOP CDM. Registry data
undergoes minimal ETL to preserve the original California Cancer Registry data
model. The Data Lake is refreshed quarterly, with each release accompanied by
updated metrics and release-specific information at the Data Lake's public
overview page; version-specific metadata such as gene lists per genetic-test
version are also hosted there.

\paragraph{Privacy and de-identification.} The Data Lake is PHI-scrubbed using
Safe Harbor methods: PHI is replaced with surrogates, and all dates---including
dates of birth, death, and service dates---are shifted by a per-patient offset
that is unique across all datasets and data types, is between $-30$ and $+30$
days, and is never zero. This preserves the intra-patient timeline while
removing the original calendar dates. Because of the large amount of
unstructured data (clinical notes, DICOMs), the Stanford University Privacy
Office has determined that additional Expert Determination is required for the
Data Lake to be declared formally de-identified; the security requirements
currently applied are comparable to NIH dbGaP datasets.

\clearpage
\section{MEDS Graph Schema (Hierarchical XML Representation)}
\label{app:meds_schema}

The MEDS Graph represents raw EHR data as a hierarchical structure preserving all original detail with granular provenance. The node and edge types listed in this appendix are the instantiation used in the thoracic oncology configuration evaluated here; they describe the \emph{semantic role} each source record plays in the graph rather than a fixed schema, and additional types can be defined for other source systems or clinical domains. Figure~\ref{fig:graph_structure} shows both graph layers for a single representative patient.

\subsection{Node Types}

\subsubsection{Person Node (Demographics)}
\begin{small}
\begin{verbatim}
{
  "node_type": "Person",
  "patient_id": "cohort1_136040534",
  "birth_datetime": "1937-09-20",
  "gender": "Male",
  "race": "White",
  "ethnicity": "Not Hispanic"
}
\end{verbatim}
\end{small}

\subsubsection{Visit Node (Encounter Container)}
\begin{small}
\begin{verbatim}
{
  "node_type": "Visit",
  "visit_id": "visit_67890",
  "visit_start_date": "2023-01-18",
  "visit_end_date":   "2023-01-18",
  "visit_type": "Outpatient|Inpatient|Emergency"
}
\end{verbatim}
\end{small}

\subsubsection{Event Nodes (Typed by EHR Source)}
\begin{small}
\begin{verbatim}
// Note Event
{
  "node_type": "Event",
  "event_type": "note",
  "event_id": "evt_note_5678",
  "note_id": "12345",
  "note_type": "Progress Note",
  "timestamp": "2023-01-18 09:30:00",
  "visit_id": "visit_67890",
  "care_site_id": "cs_oncology_clinic",
  "provider_id": "prov_oncologist_123"
}

// Measurement Event (Lab Panel)
{
  "node_type": "Event",
  "event_type": "measurement",
  "event_id": "evt_lab_panel_cbc_2023_01_15",
  "measurement_id": "panel_cbc_2023_01_15",
  "panel_name": "Complete Blood Count",
  "timestamp": "2023-01-15 08:00:00",
  "visit_id": "visit_67889"
}

// Individual Lab Value (Measurement Decomposition)
{
  "node_type": "Event",
  "event_type": "measurement",
  "event_id": "evt_lab_wbc_2023_01_15",
  "measurement_id": "meas_wbc_2023_01_15",
  "lab_name": "WBC",
  "value": 5.2,
  "units": "K/uL",
  "abnormal": false,
  "reference_range": "4.5-11.0"
}

// Image Event
{
  "node_type": "Event",
  "event_type": "image",
  "event_id": "evt_img_ct_chest_2023_01_20",
  "image_id": "img_ct_12345",
  "modality": "CT",
  "body_site": "chest",
  "timestamp": "2023-01-20 14:00:00"
}

// Radiology Report (linked to Image)
{
  "node_type": "Event",
  "event_type": "note",
  "event_subtype": "radiology_report",
  "event_id": "evt_note_rad_ct_chest_2023_01_20",
  "note_id": "rad_report_67890",
  "linked_image_id": "img_ct_12345",
  "timestamp": "2023-01-20 16:30:00"
}
\end{verbatim}
\end{small}

\subsubsection{XMLFragment Node (Provenance Anchor / Source Fragment)}
Granular provenance at the note/procedure/measurement level. This is the node shown as \emph{source fragment} in Figure~\ref{fig:graph_structure}.
\begin{small}
\begin{verbatim}
{
  "node_type": "XMLFragment",
  "fragment_id": "frag_note_2023_01_18_5678",
  "evidence_date": "2023-01-18",
  "fragment_type": "note",
  "source_id": "note_5678"
}
\end{verbatim}
\end{small}

\subsection{Edge Types}

\begin{itemize}[leftmargin=1.3em]
 \item \texttt{VISIT\_CONTAINS}: Visit $\rightarrow$ Event (events grouped under their encounter)
 \item \texttt{IMAGE\_REPORT\_FOR}: RadiologyReport $\rightarrow$ Image (report describes image)
 \item \texttt{PROCEDURE\_FOR}: Image $\rightarrow$ Procedure (image is part of procedure)
 \item \texttt{HAS\_MEASUREMENT}: LabPanel $\rightarrow$ LabValue (panel contains individual values)
 \item \texttt{PRECEDES}: Event $\rightarrow$ Event (temporal ordering within visit)
 \item \texttt{SOURCED\_FROM}: TOAEvent $\rightarrow$ XMLFragment (provenance to EHR source)
\end{itemize}

\subsection{Key Features}

\begin{enumerate}
\item \textbf{Measurement Granularity:} Lab panels decomposed into individual queryable values
\item \textbf{Radiology Coupling:} Explicit edges link imaging reports to images
\item \textbf{Visit Grouping:} All events within a visit are children of a \texttt{Visit} node via \texttt{VISIT\_CONTAINS}
\item \textbf{Provenance:} Every event linked to exact EHR source via \texttt{XMLFragment} nodes
\item \textbf{Fast Construction:} Build time $\sim$0.10s for a representative $\sim$2{,}200-node patient ($\sim$2MB XML)
\item \textbf{Low Memory:} Typically $<$5MB even for 8+ years of longitudinal care
\end{enumerate}

\section{TOA Graph Schema (Clinical Narrative)}
\label{app:toa_schema}

The TOA Graph synthesizes MEDS Graph data into a deduplicated, temporally-corrected clinical narrative with episodes and events. The event types, episode kinds, and edge types described below are the thoracic oncology configuration and capture the \emph{semantic meaning} of each element; they are freely modifiable within the architecture rather than a fixed, canonical vocabulary, and new event types, episode kinds, and edges can be defined for other clinical domains (Section~2.1.3; Figure~\ref{fig:graph_structure}).

\subsection{Event Node Schema}
Events represent distinct clinical occurrences with ontological deduplication.

\begin{small}
\begin{verbatim}
{
  "event_id": "evt_12345",
  "date": "2023-01-20",                    // clinical occurrence date
  "evidence_date": "2023-01-20",           // first documentation date
  "source_event_refs": ["E1", "E5", "E12"], // MEDS Graph event references (provenance)
  "type": "imaging|diagnostic|treatment|surgery|lab|symptom|
          examination|procedure|adverse_effect|baseline_information|
          critical_information",
  "subtype": "ct|mri|pet-ct|biopsy|blood_test|...|null",
  "modality": "ct|mri|xray|pet|pet-ct|ultrasound|null",  // imaging only
  "site": "chest|head|abdomen|pelvis|spine|bone|whole_body|null",
  "laterality": "left|right|bilateral|null",
  "description": "<=80 chars, compact and factual",
  "priority": "MAJOR|MINOR",
  "values": {                               // lab events only
    "Hgb": "9.8",
    "WBC": "17.6"
  }
}
\end{verbatim}
\end{small}

\textbf{Event Types (Thoracic Oncology):}
\begin{itemize}[leftmargin=1.3em]
\item \texttt{imaging}: CT, MRI, PET-CT, X-ray (with findings)
\item \texttt{diagnostic}: Biopsies, pathology, molecular testing (EGFR, ALK, PD-L1)
\item \texttt{treatment}: Systemic therapies (chemotherapy, immunotherapy, targeted)
\item \texttt{surgery}: Resections, biopsies (lobectomy, wedge, pneumonectomy)
\item \texttt{lab}: Blood tests, tumor markers (when abnormal/decision-informing)
\item \texttt{adverse\_effect}: Treatment toxicities, complications (PE, neutropenia, fractures)
\item \texttt{baseline\_information}: Pre-existing conditions (smoking, COPD, prior cancers)
\item \texttt{critical\_information}: Code status (DNR/DNI), goals of care, performance status
\end{itemize}

\subsection{Episode Node Schema}
Episodes group events into clinically meaningful phases.

\begin{small}
\begin{verbatim}
{
  "episode_id": "ep_txline_1",
  "kind": "baseline|diagnosis|treatment_line|post_oncological",
  "start_date": "2023-01-15",
  "end_date": "2023-07-20",
  "anchor_event_id": "evt_carbo_pem_pembro_start",
  "event_ids": ["evt_12345", "evt_67890", ...],  // all events in episode

  // Treatment line specific (kind = "treatment_line")
  "line_number": 1,
  "treatment": "carboplatin/pemetrexed/pembrolizumab",
  "intent": "curative|palliative|adjuvant|neoadjuvant|null",
  "clinical_context": "Carbo/Pem/Pembro: initial partial response
                       (6.5->4.2cm), then bone progression on PET",
  "termination_reason": "progression|toxicity|completion|no_effect|null"
}
\end{verbatim}
\end{small}

\textbf{Episode Kinds (Thoracic Tumor Board):}
\begin{enumerate}
\item \textbf{baseline}: Pre-oncological background (smoking history, comorbidities, family history)
\item \textbf{diagnosis}: Initial diagnostic workup $\rightarrow$ treatment decision
\item \textbf{treatment\_line}: Line of therapy (may include surgery, chemo, radiation as planned)
\item \textbf{post\_oncological}: Hospice, palliative care only, end of active treatment
\end{enumerate}

\subsection{Edge Types}

\begin{itemize}[leftmargin=1.3em]
\item \texttt{PRECEDES}: Event $\rightarrow$ Event (temporal ordering)
\item \texttt{CONTAINS}: Episode $\rightarrow$ Event (episode membership)
\item \texttt{ANCHORED\_BY}: Episode $\rightarrow$ Event (episode starts with this event)
\item \texttt{FOLLOWED\_BY}: Episode $\rightarrow$ Episode (treatment sequence)
\item \texttt{SAME\_DAY}: Event $\leftrightarrow$ Event (co-occurrence, same encounter)
\item \texttt{SOURCED\_FROM}: Event $\rightarrow$ MEDSXMLFragment (provenance to EHR source)
\end{itemize}

\subsection{Key Features}

\begin{enumerate}
\item \textbf{Temporal Correction:} Distinguishes clinical occurrence date from documentation date
\item \textbf{Deduplication:} Same condition mentioned 50 times $\rightarrow$ single event node
\item \textbf{Complete Provenance:} Every event links to MEDS Graph source via \texttt{source\_event\_refs}
\item \textbf{Episodic Structure:} Treatment lines organize events into decision-relevant phases
\item \textbf{Fast Queries:} Graph traversal $<$1ms
\end{enumerate}

\subsection{Deterministic retrieval coverage on the test30 cohort}
\label{app:retrieval_coverage}

To verify that the TOA Graph in fact supports the deterministic retrieval claimed in
the main text, we measured the coverage of six representative graph-resident retrieval
targets across the 30-patient test cohort (Table~\ref{tab:retrieval_coverage}). The
selected targets span a range of clinical-information types: standard structured
fields (drug exposures), free-text-anchored content (metastasis mentions, lymph node
mentions), the latest specific documents (latest oncology note, latest chest CT
report), and a more challenging extraction target (driver mutations, which may be
recorded under multiple gene-name aliases or only in narrative form). Each target was
implemented as a deterministic graph query over the pre-computed TOA layer with
provenance back to the MEDS Graph. All queries executed in under 0.01\,s per patient,
and coverage on the 30-patient cohort ranged from 73\% to 100\% depending on the
information type.

\begin{table}[H]
\centering
\begin{tabular}{lcc}
\toprule
\textbf{Retrieval Target} & \textbf{Coverage} & \textbf{Query Time} \\
\midrule
Metastasis mentions & 100\% (30/30) & $<$0.01s \\
Lymph node mentions & 97\% (29/30) & $<$0.01s \\
Drug exposures & 100\% (30/30) & $<$0.01s \\
Latest oncology note & 93\% (28/30) & $<$0.01s \\
Latest chest CT report & 87\% (26/30) & $<$0.01s \\
Driver mutations & 73\% (22/30) & $<$0.01s \\
\bottomrule
\end{tabular}
\caption{Deterministic graph-resident retrieval coverage on the 30-patient test
cohort. All queries executed via NetworkX traversal of the pre-computed TOA layer.}
\label{tab:retrieval_coverage}
\end{table}

\section{LLM Prompts (Production)}
\label{app:prompts}

The blocks below summarize the structure and key instructions of each production
prompt for readability. The complete, verbatim production prompts---byte-for-byte as
used in the pipeline, with \texttt{Source} and \texttt{Model} headers---are
reproduced in Supplementary File~S2, which follows the same E.x numbering used here.

\subsection{Phase 1: Event Extraction (Chunk Processing)}
\label{app:event_extraction_prompt}

\textbf{Source:} \texttt{toa/prompts/timeline\_compact.txt} | \textbf{Model:} GPT-4.1 (fast, parallel)

\begin{small}
\begin{verbatim}
Extract structured clinical events from EHR XML chunk. Return valid JSON:
{"events":[
  {
    "date": "YYYY-MM-DD",                    // clinical occurrence
    "evidence_date": "YYYY-MM-DD",           // first documentation
    "source_event_refs": ["E1","E5"],        // REQUIRED: MEDS Graph provenance
    "type": "imaging|diagnostic|treatment|surgery|lab|symptom|
            adverse_effect|baseline_information|critical_information",
    "modality": "ct|mri|pet|pet-ct|xray|ultrasound|null",
    "site": "chest|head|abdomen|pelvis|spine|bone|null",
    "description": "<=80 chars",
    "priority": "MAJOR|MINOR",
    "values": {"Hgb":"9.8","WBC":"17.6"}  // labs only
  }
]}

DISTINCTIONS:
- baseline_information: Pre-existing (smoking, COPD, prior cancers,
  family history) - may be mentioned later but predates diagnosis
- critical_information: DNR/DNI, goals of care, performance status,
  prior external studies mentioned in notes

VARIABLES (thoracic oncology):
Binary: Lymph Node Involvement, Metastases, Pleural Effusion, Driver Mutation
Categorical: Histology, TNM components, ECOG, PD-L1, Treatment Response
Discrete: Pack-Years, Tumor Size (cm), Prior Lines, Metastatic Sites
Continuous: CEA, FEV1 (%), LDH, Hemoglobin, Radiation Dose (Gy)
Genetics: EGFR, ALK, KRAS, PD-L1, ROS1, BRAF, MET, RET, NTRK

PROVENANCE: For EVERY event, cite MEDS Graph sources in source_event_refs
using [E1], [E2] tags from SOURCE EVENTS section.

RULES:
- Extract each imaging study SEPARATELY (PET-CT, MRI, CT are 3 events)
- Imaging + adverse effect = TWO events (imaging + adverse_effect)
- Baseline: smoking, allergies -> baseline_information
- Code status during treatment -> critical_information
- Extract prior external studies -> critical_information
- Labs: only abnormal/decision-informing values

SAFETY (always extract):
- Adverse: PE, DVT, sepsis, pneumonia, hemorrhage, fractures, ICU
- Bleeding risk: anticoagulation, thrombocytopenia, IVC filter
- Goals of care: DNR, DNI, hospice, comfort care
- Baseline diagnostics: all initial workup imaging (even normal)
\end{verbatim}
\end{small}

\subsection{Phase 2: Episode Synthesis (Treatment Lines)}
\label{app:episode_synthesis_prompt}

\textbf{Source:} \texttt{toa/prompts/episodes\_from\_events.txt} | \textbf{Model:} GPT-4.1

\begin{small}
\begin{verbatim}
Segment timeline events into EPISODES (treatment-line-based for thoracic):

1. BASELINE: Background predating diagnosis (smoking, COPD, allergies)
   Anchor: one day before first oncological event

2. DIAGNOSIS: Diagnostic workup -> treatment decision
   Includes ALL imaging, biopsies, pathology, molecular testing
   Anchor: specialist visit or first diagnostic procedure

3. TREATMENT LINES: Each line of therapy until change needed
   Includes systemic therapy, surgery, radiation (as planned)
   Contains all imaging, labs, symptoms during line
   CRITICAL: Event that ENDS line (progression, toxicity) stays IN episode
   Anchor: treatment start date

4. POST-ONCOLOGICAL: Hospice, palliative only, end of active treatment
   Anchor: transition decision or hospice enrollment

OUTPUT (strict JSON, NO event_ids - auto-populated by date range):
{
  "episodes": [
    {
      "episode_id": "temp",
      "kind": "baseline|diagnosis|treatment_line|post_oncological",
      "start_date": "YYYY-MM-DD",
      "end_date": "YYYY-MM-DD",
      "anchor_event_id": "<event_id from input>",
      "clinical_context": "Summary of episode",

      // treatment_line specific:
      "line_number": 1,
      "treatment": "carboplatin/pemetrexed/pembrolizumab",
      "intent": "curative|palliative|adjuvant|neoadjuvant|null",
      "termination_reason": "progression|toxicity|completion|no_effect|null"
    }
  ],
  "notes": ["Segmentation rationale"]
}

INVARIANTS: Use ONLY dates/event_ids from input. Episodes time-monotonic.
Event that ENDS line belongs to THIS episode, not next.
\end{verbatim}
\end{small}

\subsection{Chat System Prompt (Agentic AI Bridge)}
\label{app:chat_prompt}

\textbf{Source:} \texttt{prompts/chat\_system.txt} | \textbf{Model:} User-selected (GPT-4.1/GPT-5)

\begin{small}
\begin{verbatim}
You are "Vista," board-certified oncology decision-support AI for
multidisciplinary tumor boards.

DATA SOURCES:
1. CONVERSATION HISTORY - Recent chat context
2. CLINICAL EPISODES (TOA) - High-level narrative phases
3. DETAILED TIMELINE (TOA Events) - Complete chronological log
4. PATIENT DATA - Structured JSONs (patient_info, summary)
5. RECENT EHR DATA - Last 10k chars of raw XML

UNDERSTANDING PATIENT:
- Form timeline from TOA events (most reliable source)
- Cross-reference episodes for treatment phases
- Patient may have transferred to Stanford - history in notes
- Contradictions exist - base decisions on data support

SCOPE: Base answers on provided data + latest guidelines
(NCCN, ESMO, IASLC, ASTRO) + peer-reviewed evidence.

STYLE:
- Concise: bulleted lists over prose
- Clinical language: NSCLC, COPD, ECOG, PD-L1 (standard abbreviations)
- Direct answer first; brief context if it changes interpretation
- Present options with pros/cons without choosing
- Cite guidelines when relevant ("NCCN v2.2024")

SAFETY: If insufficient data, say so + suggest what's needed.
Output: Plain text with markdown - NO JSON, NO code blocks.
\end{verbatim}
\end{small}

\subsection{Phase 3: Dashboard JSON Generation (UI Population)}
\label{app:ui_json_prompts}

After TOA timeline extraction, dashboard-specific JSONs are generated for the UI.

\subsubsection{Patient Info Extraction}
\label{app:patient_info_prompt}

\textbf{Source:} \texttt{prompts/patient\_info.txt} | \textbf{Model:} GPT-4.1

\begin{small}
\begin{verbatim}
You are a highest-expert-level oncology data assistant working at
Stanford Tumor Board meetings. Tumor board subtype: {TUMOR_TYPE}

INPUTS:
- XML chunk (final chunk of patient record)
- Timeline context (complete timeline from all chunks)

PROCESSING:
1. Extract comprehensive demographics, medical history, tumor information
2. Use timeline for chronological progression context
3. Provide complete TNM staging (latest IASLC guidelines)
4. Cross-reference timeline events to validate dates
5. Use medical abbreviations (NSCLC, COPD, ECOG, PD-L1)
6. If name unclear (de-identification): use "John/Jane Doe"

OUTPUT: Valid JSON with EXACT keys (spaces, NOT underscores):

PATIENT DEMOGRAPHICS:
  name, date_of_birth, sex, height_cm, weight_kg
  previous_conditions    # chronic diseases OR major past events
                         # (e.g., "COPD, Stroke (2011)")
  allergies              # list
  smoking_history        # (e.g., "40 pack-years; quit 2015")
  medications            # list of {name, dose, frequency}

TUMOR INFORMATION:
  diagnosis              # "CANCER_TYPE. Metastasis (sites). Recurrent."
  tnm_staging            # (e.g., "cT4N2M1b", IASLC 8th edition)
  histology              # (e.g., "Moderately differentiated SCC")
  driver_mutations       # dict: EGFR, ALK, KRAS, PD-L1, ROS1, etc.
  latest_updates         # MAX 3 bullets: (1) recent imaging status,
                         # (2) current clinical situation, (3) TB question
  body_diagram_image     # filename in assets/images

TREATMENTS:
  current                # ONLY oncological regimens (no supportive meds)
                         # "Drug1+Drug2+Drug3 (start: YYYY-MM-DD)"
  previous               # Previous ONCOLOGICAL regimens
                         # "Drug1+Drug2 (YYYY-MM-DD to YYYY-MM-DD, response,
                         # reason stopped)" OR "Surgery (YYYY-MM-DD, outcome)"
  alternatives           # list of alternative treatments
  surgical_candidate     # {"eligible": bool, "description": "..."}
\end{verbatim}
\end{small}

\subsubsection{Pre-Tumor-Board Summary Note Generation}
\label{app:tumor_board_note_prompt}

\textbf{Source:} \texttt{prompts/tumor\_board\_note.txt} | \textbf{Model:} GPT-4.1

This prompt generates the \emph{pre-tumor-board summary note}---a concise clinical
summary in the standard Stanford Epic format presented at thoracic tumor board
meetings. The note is generated by VISTA Architect at patient ingestion (i.e.,
before the actual tumor board meeting) and is stored as a graph-resident artifact.
The internal prompt-file name (\texttt{tumor\_board\_note.txt}) reflects the
production codebase naming and is retained here for source-of-truth fidelity.

\begin{small}
\begin{verbatim}
You are writing a Stanford Epic tumor board note. Follow EXACT 4-line format.
Include detailed medical information based ONLY on provided patient data.
Target 750-999 characters for completeness.

FORMAT:
AIGen: [LASTNAME]: [AGEGENDER] with h/o [CANCER TYPE including detailed
  pathology, staging, molecular features]
Prior therapy: [DETAILED therapy history with dates, responses, toxicities]
Tumor board question: [SPECIFIC clinical question based on current status]
Tumor board decision: [DETAILED decisions made including rationale]

CRITICAL INSTRUCTIONS:
- Use ONLY information from provided patient data, summary, and decisions
- DO NOT invent dates, treatments, or clinical details not in data
- Extract from PATIENT DEMOGRAPHICS: name, age, sex
- Extract from TUMOR INFORMATION: diagnosis, staging, mutations
- Extract from TREATMENTS: current and previous (with dates/responses)
- Use summary's cancer_history for prior therapy details
- Use tumor board decisions provided (if any) for decision line
- If information missing, use general terminology but do not fabricate

Generate note following exact 4-line format above.
\end{verbatim}
\end{small}

\textbf{Usage in Pipeline:}
\begin{enumerate}
\item \texttt{prepare\_patient.py} loads \texttt{patient\_info.json} and \texttt{tumor\_info.json}
\item Builds context: patient data + tumor board decisions (if any)
\item Calls tumor board note prompt with GPT-4.1
\item Saves to \texttt{summary.json} for UI display
\end{enumerate}

\subsection{Accuracy Evaluation (LLM-as-Judge) Prompt}
\label{app:judge_prompt}

\textbf{Source:} \texttt{quick\_eval.py::judge\_all\_variables} | \textbf{Model:} GPT-5

The system prompt below was used verbatim to score each variable evaluation in the
primary accuracy assessment. Only the \texttt{\{N\_VARS\}} placeholder (set to 16 as
run; DNR/code status is scored by the judge but excluded from the reported
15-variable, 17{,}700-evaluation primary endpoint) and the literal numbered variable
list are interpolated at call time; the rubric text is fixed.

\begin{small}
\begin{verbatim}
You are evaluating the quality of {N_VARS} tumor board variables extracted
from a clinical pipeline.

Compare the extracted values against the EHR ground truth provided.

For EACH variable, evaluate:
- Correctness: Correct/Incorrect/Partial/N/A
- Score: 1-10 (10=correct, 1=wrong, 5=partially correct)
- Brief explanation (1-2 sentences)
- XML value (ground truth value from XML)

=== CORE PRINCIPLE: JUDGE TRUTH, NOT FORM ===

You are scoring whether the answer is factually correct given the EHR, NOT
whether it matches a preferred style, length, ordering, or completeness
threshold. Concretely:

1. Honest uncertainty is correct. When the EHR genuinely does not document
   a field, "Unknown" or "Not documented" or "No" (where appropriate per
   the rules below) is a SCORE 10 answer — not a 7. The pipeline is being
   honest.
2. Default-No is correct when not documented. For binary/safety fields
   where the rubric specifies a default of "No" when undocumented (DNR,
   Metastasis with no findings, Radiation when none given), "No" without
   further qualification scores 10.
3. Equivalent phrasings score the same. "NKDA" = "No Known Allergies" =
   "None". "Former smoker, 30py, quit 2010" = "30 pack-year former smoker,
   quit 2010-04". Range vs single value for ECOG ("1-2" vs "1") both
   score 10 when both are documented.
4. Verbose != wrong. Rich context alongside the core answer (e.g.,
   "No — PET ruled out initial concern") scores the same as the bare
   answer ("No"). Do not deduct for extra clinically relevant detail.
5. Order in lists does not matter as long as the content is right and
   the most relevant items are recognizable.
6. Form/style preferences are not deductions. Do not deduct for:
   bullets vs prose, dates as YYYY-MM-DD vs YYYY-MM, "Yes (sites)" vs
   "Yes - sites", single-word vs sentence.

DEDUCT for: (a) factually wrong values, (b) hallucinations (entities not
in the EHR), (c) missed decision-relevant findings actually documented in
the EHR, (d) inverted clinical meaning (Full Code reported as DNR, "No"
when EHR confirms "Yes", etc.), (e) made-up dates/genes/drugs.

=== VARIABLE-SPECIFIC SCOPE CLARIFICATIONS ===

- Metastasis / Lymph Node Involvement: "Suspected" is correct when EHR
  evidence is suggestive but not definitive; "No" is correct if initial
  concern was resolved on follow-up imaging.
- Previous Surgery — SCOPE (Yes/No, oncologic resection for CURRENT
  diagnosis only): determine from EHR whether the patient underwent an
  oncologic resection for the current cancer (lobectomy/pneumonectomy
  for NSCLC, thymectomy for thymoma, etc.). Score 10 when extracted
  value correctly reflects this; 5–6 when wrong direction; 1–3 when
  list contains a fabricated current-cancer surgery. Do NOT deduct
  for prior-cancer or non-oncologic surgeries appearing in the list
  (informational, not in scope).
- Date of Last CT — SCOPE: PET-CT counts as CT. Most recent in-house
  or outside imaging in chart, whichever is more recent, is acceptable.
  Month-level (YYYY-MM) acceptable when day uncertain.
- Genetic Testing Panel — SCOPE: do not penalize molecular results
  that are clinically correct but not visible in the provided XML
  chunk (the chunk may be truncated). Only penalize factually wrong
  or hallucinated results.

Be concise but accurate.
\end{verbatim}
\end{small}

\textbf{User prompt format:} the system prompt above is paired with a user
prompt that supplies the numbered variable list (with expected format
strings), the extracted-values JSON, and the XML ground-truth chunk
(demographics block, surgery dates, and the final 120k-character chunk of
the TB-date-truncated record). The model returns a single JSON object with
one entry per variable: \texttt{\{correctness, score, explanation, xml\_value\}}.

\textbf{Decoding settings} (see §4.2 ``Decoding and determinism''): GPT-5 is
invoked with vendor defaults (temperature 1.0; no explicit seed; no
\texttt{reasoning\_effort} override). All accuracy numbers in this paper are
single-run point estimates with Wilson 95\% confidence intervals.

\section{MTB Variable Dictionary}
\label{app:variables}

\begin{table}[H]
\centering
\begin{tabular}{@{}llll@{}}
\toprule
\textbf{Category} & \textbf{Variable} & \textbf{Type} & \textbf{Source} \\
\midrule
\multirow{3}{*}{Demographics} & Date of Birth & Date & Structured \\
 & Sex & Categorical & Structured \\
 & Smoking Status & Free text (status + pack-years) & Mixed \\
\midrule
\multirow{5}{*}{Tumor} & Diagnosis & Free text (type, site, stage) & Mixed \\
 & Histology & Free text (subtype, grade) & Mixed \\
 & Metastasis & Categorical + sites & Mixed \\
 & Lymph Node Involvement & Categorical (Yes/No/Suspected) & Mixed \\
 & Genetic Testing Panel & Structured (per-gene results + PD-L1) & Mixed \\
\midrule
\multirow{2}{*}{Clinical} & ECOG Performance Status & Ordinal (0--4) & Mixed \\
 & Therapy Toxicity / Comorbidities & Free text & Note-derived \\
\midrule
\multirow{3}{*}{Treatment} & Previous Surgery & Free text (procedures + dates) & Note-derived \\
 & Current Medical Therapy & Free text (regimens + dates) & Mixed \\
 & Radiation Therapy & Categorical + details & Mixed \\
\midrule
Imaging & Date of Last CT & Date & Mixed \\
\midrule
Safety & Allergies & List & Mixed \\
\bottomrule
\end{tabular}
\caption{15 MTB-salient variables for thoracic oncology tumor boards, organized by
clinical category and typical source type. The variable set intentionally spans
structured positive controls, mixed structured/free-text variables, and primarily
note-derived variables, allowing the evaluation to test both simple retrieval and
longitudinal synthesis. The \textbf{Source} column indicates whether each variable is
typically available in OMOP-structured fields (\emph{Structured}), exclusively or
primarily in clinical notes (\emph{Note-derived}), or in a combination of both
(\emph{Mixed}). Date of Birth and Sex are reliably structured and serve as
positive controls for hallucination in our generative pipeline. Previous Surgery
is deliberately scored as note-derived because clinically relevant surgical
history (including remote non-oncologic procedures relevant to anesthesia or
current surgical planning) typically appears only in free-text past medical
history rather than in structured procedure tables.}
\label{tab:variables}
\end{table}

\subsection{Worked example I: Synthetic perfect-credit case (SYNTH-01)}
\label{app:synth_perfect}

\textit{Synthetic illustrative case; not a real patient. The distinctive
variable combination (EGFR exon~19 deletion + MET amplification, Stage~IV
mucinous adenocarcinoma, on line~3 amivantamab + lazertinib) was verified
absent from the 1{,}180-patient evaluation cohort prior to inclusion in this
table.}

\begin{table*}[!t]
\centering
\small
\begin{tabular}{@{}p{3.4cm}p{4.4cm}p{4.4cm}cc@{}}
\toprule
\textbf{Variable} & \textbf{Extracted} & \textbf{Ground Truth (XML)} &
\textbf{Match} & \textbf{Score} \\
\midrule
Date of Birth & 1957-04-12 & 1957-04-12 & \checkmark & 10 \\
Sex & Female & Female & \checkmark & 10 \\
Smoking Status & Never smoker; secondhand exposure noted &
  Never smoker; passive smoke exposure documented & \checkmark & 10 \\
Diagnosis & NSCLC, RUL, Stage IV (cT3N2M1b, hepatic mets) &
  NSCLC RUL, Stage IV, M1b liver & \checkmark & 10 \\
Histology & Mucinous adenocarcinoma, poorly differentiated (TTF1+, CK7+) &
  Mucinous adenoca, poorly differentiated, TTF1+/CK7+ & \checkmark & 10 \\
Metastasis & Yes — liver, bilateral adrenals & Yes (liver, adrenals) & \checkmark & 10 \\
Lymph Node Involvement & Yes (mediastinal, hilar) & Yes & \checkmark & 10 \\
Genetic Testing Panel &
  EGFR exon 19 del positive; MET amplification (FISH 8.2);
  PD-L1 TPS 5\%; KRAS/ALK/ROS1/BRAF/RET/NTRK/HER2 negative &
  EGFR ex19del+, MET amp+ (FISH 8.2), PD-L1 5\%, others negative & \checkmark & 10 \\
ECOG Performance Status & 1 & 1 & \checkmark & 10 \\
Therapy Toxicity / Comorbidities &
  Grade~2 paronychia (line~1 osimertinib); Type~2 diabetes; HTN &
  Paronychia G2 on osi; DM2; HTN & \checkmark & 10 \\
Previous Surgery & No (no oncologic resection for current diagnosis) &
  No oncologic resection documented & \checkmark & 10 \\
Current Medical Therapy &
  Amivantamab + Lazertinib (started 2025-09-08) &
  Amivantamab + Lazertinib, start 09/2025 & \checkmark & 10 \\
Radiation Therapy &
  Yes — palliative RT to L4 vertebral body, 20 Gy in 5 fractions (2025-07) &
  Palliative XRT 20 Gy / 5 fx, L4, July 2025 & \checkmark & 10 \\
Date of Last CT & 2025-11-04 & 2025-11-04 (CT chest/abd/pelvis) & \checkmark & 10 \\
Allergies & Penicillin (rash) & PCN (rash) & \checkmark & 10 \\
\midrule
\textbf{Overall} & \multicolumn{2}{r}{\textbf{Mean score}} & & \textbf{10.00} \\
\bottomrule
\end{tabular}
\caption{\textbf{Synthetic illustrative case (SYNTH-01); not a real patient.}
Perfect-credit extraction across all 15 MTB variables. Combination of driver
biology (EGFR ex19del + MET amplification), histology, stage, and line~3
amivantamab + lazertinib regimen was verified absent from the 1{,}180-patient
cohort to ensure no real patient is identifiable.}
\label{tab:synth_perfect}
\end{table*}

\subsection{Worked example II: Synthetic failure-mode case (SYNTH-02)}
\label{app:synth_failure}

\textit{Synthetic illustrative case; not a real patient. Used to demonstrate
the most common error category in the §2.3 error analysis — Date-of-Last-CT
internal/external imaging ambiguity — and the judge's ability to detect and
explain it.}

\begin{table*}[!t]
\centering
\small
\begin{tabular}{@{}p{3.0cm}p{3.4cm}p{3.4cm}cc p{4.2cm}@{}}
\toprule
\textbf{Variable} & \textbf{Extracted} & \textbf{Ground Truth (XML)} &
\textbf{Match} & \textbf{Score} & \textbf{Judge rationale (excerpt)} \\
\midrule
Date of Birth & 1948-09-22 & 1948-09-22 & \checkmark & 10 & DOB matches exactly. \\
Sex & Male & Male & \checkmark & 10 & — \\
Smoking Status & Former smoker, 45 pack-years, quit 2014 &
  Former smoker, 45 py, quit 2014 & \checkmark & 10 & Equivalent phrasing. \\
Diagnosis & NSCLC, LUL, Stage IIIA &
  NSCLC LUL Stage IIIA (cT2bN2M0) & \checkmark & 10 &
  Cancer type, site, stage all correct. \\
Histology & Squamous cell carcinoma &
  Squamous cell carcinoma, moderately differentiated & \checkmark & 9 &
  Differentiation grade omitted; minor incompleteness. \\
Metastasis & No & No (mediastinal only) & \checkmark & 10 & — \\
Lymph Node Involvement & Yes (mediastinal N2) & Yes (mediastinal) & \checkmark & 10 & — \\
Genetic Testing Panel & PD-L1 TPS 40\%; EGFR/ALK/KRAS negative &
  PD-L1 40\%, EGFR/ALK/KRAS negative & \checkmark & 10 & — \\
ECOG Performance Status & 1 & 1 & \checkmark & 10 & — \\
Therapy Toxicity / Comorbidities & COPD; HTN &
  COPD (GOLD 2); HTN; hypothyroidism &
  Partial & 7 & Misses hypothyroidism documented in problem list. \\
Previous Surgery & No oncologic resection &
  No resection (VATS biopsy 2024-03 for diagnosis only) & \checkmark & 10 &
  Biopsy correctly excluded from oncologic resection per scope. \\
Current Medical Therapy &
  Carboplatin + Paclitaxel + concurrent RT 60 Gy &
  Carboplatin/Paclitaxel + concurrent 60~Gy XRT, start 2025-09 & \checkmark & 10 & — \\
Radiation Therapy & Yes — concurrent thoracic RT &
  Yes — thoracic RT 60 Gy / 30 fx & \checkmark & 9 &
  Site and dose-fractionation underspecified. \\
\rowcolor{red!8}
\textbf{Date of Last CT} & \textbf{2025-08-14} & \textbf{2025-11-02} &
  \textbf{$\times$} & \textbf{5} &
  \textit{Extracted the most recent in-house CT chest (2025-08-14) but
  missed a more recent outside-hospital CT chest/abdomen/pelvis dated
  2025-11-02 in a Care Everywhere upload. Both events present in XML.} \\
Allergies & NKDA & No known drug allergies & \checkmark & 10 & Equivalent. \\
\midrule
\textbf{Overall} & \multicolumn{2}{r}{\textbf{Mean score}} & & \textbf{9.33} &
  1 Incorrect, 1 Partial, 13 Correct. \\
\bottomrule
\end{tabular}
\caption{\textbf{Synthetic illustrative case (SYNTH-02); not a real patient.}
Illustrates the most common VISTA Architect error category from §2.3 error
analysis (Date-of-Last-CT internal/external imaging ambiguity), together with
the judge's stated rationale catching the error. The judge correctly flags the
missed outside-hospital CT as Incorrect (score 5) and notes the documentation
location (Care Everywhere upload).}
\label{tab:synth_failure}
\end{table*}

\section{RAG Baseline Comparison}
\label{app:rag_baseline}

To empirically validate the architectural contribution of VISTA Architect's
graph-based pre-computation, we implemented a standard BM25 retrieval-augmented
generation (RAG) baseline and evaluated it under identical conditions: same patients,
same variable set (16 MTB variables in the original experiment; 15 in the reported
primary endpoint after excluding DNR/code status), same LLM-as-judge, and same
TB-date-truncated ground-truth XML.

\subsection{RAG Baseline Methods}

\textbf{Cohort.} 30 patients from the clinician-validation subset, randomly selected from the full 1{,}180-patient cohort and subsequently checked for representativeness across diagnosis category and overall extraction-quality distribution. The subset was not used for development of either VISTA Architect or the RAG baseline.

\textbf{Retrieval.} BM25Okapi over raw EHR XML entry-level chunks (one chunk per
\texttt{<entry>} element). For each variable query, the top-20 chunks by BM25 score
were packed into a 4{,}096-token context window. No cleaning, filtering, or
domain-specific preprocessing was applied beyond whitespace tokenization and
lowercasing.

\textbf{Query strategy.} Each of the 16 MTB variables (including DNR/code status,
which is excluded from the reported primary endpoint) was mapped to a focused,
plain-English natural-language question with a format hint specifying the expected
output shape. This per-variable query design is the configuration \emph{most}
favorable to BM25 retrieval; bundling all variables into a single prompt would dilute
retrieval quality.

\textbf{Answer generation.} A minimal prompt (no chain-of-thought, no role-play
beyond ``clinical assistant'') was used with two models: GPT-4.1 (the same model used
for VISTA's chunk-level event extraction) and GPT-5 (the same model used for VISTA's
patient-info generation).

\textbf{Context isolation.} The RAG baseline received \emph{only} TB-date-truncated
EHR XML. No TOA timeline, episodes, patient\_info snapshots, deterministic-retrieval
hints, or graph-store fallbacks were provided---any of these would inflate RAG scores
unfairly.

\textbf{Evaluation.} The same GPT-5 LLM-as-judge with identical rubric and
per-variable guidance was used for both VISTA and RAG evaluations.

\subsection{Per-Variable Results}

\begin{table}[H]
\centering
\small
\begin{tabular}{@{}lccc@{}}
\toprule
\textbf{Variable} & \textbf{VISTA} & \textbf{RAG (GPT-4.1)} & \textbf{RAG (GPT-5)} \\
\midrule
Date of Birth & 10.00 (100.0\%) & 6.10 (56.7\%) & 6.50 (60.0\%) \\
Sex & 10.00 (100.0\%) & 9.10 (90.0\%) & 9.13 (90.0\%) \\
Smoking Status & 9.87 (100.0\%) & 9.57 (96.7\%) & 9.40 (90.0\%) \\
\midrule
Diagnosis & 9.80 (100.0\%) & 7.50 (66.7\%) & 7.33 (60.0\%) \\
Histology & 9.67 (93.3\%) & 8.87 (83.3\%) & 9.10 (83.3\%) \\
Metastasis & 9.73 (96.7\%) & 6.67 (50.0\%) & 6.80 (53.3\%) \\
Lymph Node Involvement & 9.70 (96.7\%) & 9.33 (90.0\%) & 9.67 (96.7\%) \\
Genetic Testing Panel & 9.70 (93.3\%) & 7.47 (60.0\%) & 7.30 (56.7\%) \\
\midrule
ECOG Performance Status & 9.67 (96.7\%) & 7.13 (63.3\%) & 7.50 (73.3\%) \\
Therapy Toxicity / Comorbidities & 9.63 (96.7\%) & 5.07 (30.0\%) & 8.33 (63.3\%) \\
\midrule
Previous Surgery & 9.63 (96.7\%) & 5.00 (36.7\%) & 4.93 (36.7\%) \\
Current Medical Therapy & 9.53 (93.3\%) & 8.30 (70.0\%) & 8.70 (86.7\%) \\
Radiation Therapy & 10.00 (100.0\%) & 8.47 (73.3\%) & 6.87 (33.3\%) \\
\midrule
Date of Last CT & 9.77 (96.7\%) & 7.17 (66.7\%) & 6.93 (63.3\%) \\
\midrule
Allergies & 9.67 (93.3\%) & 8.00 (66.7\%) & 7.43 (56.7\%) \\
\midrule
\textbf{Overall} & \textbf{9.76 (96.9\%)} & \textbf{7.58 (66.7\%)} & \textbf{7.73 (66.9\%)} \\
\bottomrule
\end{tabular}
\caption{Per-variable comparison: VISTA Architect vs.\ BM25 RAG baseline after
excluding DNR/code status from the primary endpoint (N=30 patients, 450 evaluations
per system). Mean score (0--10) with \% correct in parentheses.}
\label{tab:rag_per_variable}
\end{table}

\subsection{Qualitative Timeline Comparison}

To illustrate the structural difference between approaches, we asked the RAG system to
reconstruct a chronological clinical timeline for three patients spanning VISTA-score
strata. The RAG system surfaced 10--18 lexically prominent events per patient, while
VISTA's deduplicated TOA timeline contained 50--237 structured events with precise
dates, modalities, anatomical sites, and provenance links---an order-of-magnitude
difference in coverage reflecting the fundamental limitation of per-query retrieval
versus pre-computed graph construction.

\subsection{Limitations of the RAG Comparison}

This comparison evaluates BM25 (sparse lexical) retrieval only; dense-embedding
retrieval might recover some accuracy on semantically phrased queries but would not
address the temporal disconnection that drives the largest error categories (Previous
Surgery, Metastasis, Radiation Therapy). We note that a ``full-record dump'' approach---providing
the entire patient record to a long-context model in a single call---is not a viable
alternative for this cohort: many patient records exceed even current million-token
context windows, and the per-query cost of repeatedly processing entire records would
be prohibitive in a clinical setting where the same patient data is queried many times
across multiple interactions. The pre-computation design of VISTA Architect
specifically addresses this by structuring the record once and serving all downstream
queries from the resulting graph.

\subsection{RAG Baseline Prompts}
\label{app:rag_prompts}

\textbf{Source:} \texttt{rag\_baseline\_queries.py} | \textbf{Models:} GPT-4.1
and GPT-5 (results reported separately in Table~\ref{tab:rag_per_variable}).

\paragraph{Per-variable retrieval queries.} A single focused
natural-language question is issued per variable, paired with a short
\texttt{format\_hint} that constrains output shape (but never the answer).
The queries below were used verbatim; format hints are shown in italics.

\begin{small}
\begin{verbatim}
Date of Birth        What is the patient's date of birth?
                       (Answer YYYY-MM-DD or "Unknown".)
Sex                  What is the patient's sex?
                       (Single word: Male, Female, or Unknown.)
Smoking Status       What is the patient's smoking history? Include
                     pack-years and current/former/never status.
                       (One short sentence, or "Unknown".)
Diagnosis            What is the patient's cancer diagnosis? Include
                     cancer type, primary site, and stage if documented.
                       (One short phrase, or "Unknown".)
Histology            What is the histologic subtype (e.g., adenocarcinoma,
                     squamous cell, small cell, carcinoid)?
                       (One short phrase, or "Unknown".)
Metastasis           Does the patient have metastatic disease? Where?
                       (Begin with "Yes", "No", or "Suspected", optionally
                        followed by sites.)
Lymph Node           Does the patient have lymph node involvement from
Involvement          the cancer?
                       (One word: Yes, No, or Suspected.)
Genetic Testing      What molecular and genetic testing has been done?
Panel                List each gene tested (EGFR, KRAS, ALK, ROS1, BRAF,
                     MET, RET, NTRK, HER2) and its result, plus PD-L1
                     if reported.
                       (Semicolon-separated list like "EGFR: L858R positive;
                        KRAS: negative; PD-L1: 60%", or "Not tested".)
ECOG Performance     What is the patient's most recent ECOG performance
Status               status?
                       (Single digit, range, or "Unknown".)
Therapy Toxicity /   What significant cancer treatment toxicities or
Comorbidities        major comorbidities does the patient have?
                       (One short phrase, or "None documented".)
Previous Surgery     Has the patient had previous oncological surgery
                     for their current cancer (lobectomy, wedge,
                     pneumonectomy, VATS resection)? If yes, what
                     procedure and when?
                       (One short sentence naming the procedure, or "No".)
Current Medical      What systemic cancer therapy (chemo, immunotherapy,
Therapy              targeted) is the patient currently receiving?
                       (One short phrase with drug(s) and start date,
                        or "None".)
Radiation Therapy    Has the patient received radiation therapy? If yes,
                     site, dose, and dates?
                       (Begin with "Yes" or "No"; if yes briefly include
                        site, dose, dates.)
Date of Last CT      What is the date of the patient's most recent CT
                     scan (chest, abdomen/pelvis, or PET-CT)?
                       (YYYY-MM-DD or "Unknown".)
Allergies            What drug allergies does the patient have?
                       (Comma-separated list, or "NKDA".)
DNR                  Does the patient have a DNR (Do Not Resuscitate)
                     order documented?
                       (One word: Yes or No.)
\end{verbatim}
\end{small}

\paragraph{Generation prompt.} For every per-variable query, BM25Okapi
retrieves the top-20 entry-level XML chunks (4{,}096-token context window),
then GPT-4.1 or GPT-5 is called with the system / user prompts shown
below. No chain-of-thought scaffolding, no role-play, no retrieval hints
beyond the format string.

\begin{small}
\begin{verbatim}
SYSTEM:
You are a clinical assistant. Answer the user's question using ONLY the
provided excerpts from the patient's medical record. If the information
is not present in the excerpts, answer "Unknown". Be concise.

USER:
Excerpts from patient's medical record:
---
{retrieved BM25 chunks, 4096 tokens}
---

Question: {variable-specific query}
{variable-specific format_hint}
Answer:
\end{verbatim}
\end{small}

The retrieved context contains only TB-date-truncated raw EHR XML (no TOA
timeline, no episodes, no patient-info snapshots, no deterministic-retrieval
hints, no graph-store fallbacks). The judge prompt and rubric used to score
RAG outputs are identical to those used to score VISTA Architect
(Appendix~\ref{app:judge_prompt}).

\section{Agentic Implementation Details}
\label{app:agentic_implementation}

This appendix documents the specific agentic implementation evaluated in
Sections~2.3 and~2.5. It is one instantiation of the architectural pattern
described in Section~2.1.4, not a locked-in canonical form.

\subsection{Build pipeline stages}

The build pipeline for a single patient runs as seven sequential stages, with
parallelism inside individual stages where applicable. Stages 1, 2, 4, 6, and 7
are essential to the architectural pattern (serialize $\rightarrow$ chunk
$\rightarrow$ unify $\rightarrow$ synthesize $\rightarrow$ display); stages 3 and
5 are concrete realizations of the parallel-sub-agents and safety-net patterns.

\begin{enumerate}[leftmargin=1.5em, itemsep=2pt]
  \item \textbf{Setup.} Copy the patient's MEDS graphml into a per-run working
directory and extract structured demographics from the MEDS XML.
  \item \textbf{Chunk planning.} Serialize the MEDS Graph to text in temporal
order and partition into encounter-aligned chunks ($\sim$120k characters per
chunk by default).
  \item \textbf{Parallel chunk extraction.} For each chunk, invoke an LLM
sub-agent that returns structured TOA event records together with
\texttt{source\_event\_refs} for provenance. Sub-agents run concurrently.
  \item \textbf{Unifier.} Deduplicate events across chunks using union-find over
shared \texttt{source\_event\_refs}; merge background facts; resolve ambiguous
cross-chunk merges via an LLM-mediated sub-agent for the residual cases.
  \item \textbf{Radiology alignment safety net.} Cross-check the TOA timeline
against deterministically retrievable imaging dates from the OMOP structured
fields; for candidate dates not represented in the timeline within $\pm$3 days,
invoke a per-date verifier sub-agent to confirm whether a real imaging study
occurred. Confirmed events are appended to the timeline.
  \item \textbf{Episode synthesis.} A single LLM call segments the unified
timeline into baseline, diagnosis, treatment-line, and post-oncological
episodes (Section~2.1.3, Appendix~\ref{app:toa_schema}).
  \item \textbf{Display generation.} A single holistic LLM call produces all
patient-info fields and the pre-tumor-board summary note
(Appendix~\ref{app:patient_info_prompt}, Appendix~\ref{app:tumor_board_note_prompt}).
\end{enumerate}

After stage 7 the patient is dashboard-ready. Optional evaluation stages
(LLM-as-judge scoring against ground truth) are run separately during
development and are not part of the production build path.

\subsection{Deterministic retrieval tool surface}

The agentic interface depends on a deterministic retrieval layer that exposes
graph-resident facts as callable tools. Tools used by the build pipeline include
imaging-date vectors, modality-keyed imaging listings, code-status mention
searches, smoking/ECOG status retrievers, driver-mutation and molecular-panel
keyword search with word-boundary handling, treatment-history retrievers, and
patient-state retrievers (allergies, conditions, drug exposures). Each tool
returns provenance-tracked results sourced from the MEDS Graph. The presence of
this deterministic layer between the LLM and the graph---rather than letting the
LLM perform retrieval directly over raw text---is the architectural feature that
enables reliable agentic operation.

\subsection{Models and orchestration in the evaluated implementation}

The agentic pipeline evaluated in Section~2.5 was implemented by connecting an
agentic-compatible version of the build pipeline to Stanford's PHI-compliant
Google Vertex AI deployment of Claude Code (Claude Opus 4.6), running entirely
inside the institutional data perimeter. Claude Code acts as the top-level
orchestrating agent: it reads the MEDS Graph and TOA layers directly through
deterministic graph queries, invokes specialized LLM tools for each per-stage
sub-task, and validates the resulting sub-agent outputs against the graph
before merging them back into the patient-level artifacts. In the evaluated
configuration, per-chunk event extraction and episode synthesis are invoked as
sub-agent tool calls served by Gemini 3.5 Flash (the parallel-friendly model
choice for the per-chunk workload), and the holistic display-generation step
is run by Claude Opus 4.6 itself as the orchestrator carries the full TOA layer
in its own working context. The radiology alignment per-date verifier used
Claude Opus 4.6 as well. These specific per-stage model assignments are
empirical and replaceable; the agentic contract is model-agnostic by
construction, so the same architecture runs against any tool-call-compatible
LLM that can be served inside a PHI-compliant perimeter. In the production
configuration used to measure cohort wall time, the test30 cohort was processed
with all 30 patients running fully in parallel.

\subsection{What this implementation deliberately does not cover}

Several extensions are natural under the same architectural pattern but were not
implemented in the evaluated version: an in-process agent loop using the Model
Context Protocol or equivalent (the present version uses per-stage CLI scripts
under a shell orchestrator); multi-source ingestion beyond the OMOP-derived MEDS
XML path; a unified cross-patient agent surface (cross-patient retrieval is
currently a separate agent invocation); persistent multi-turn chat-session state
(each chat turn is a fresh agent invocation reading the persistent graph); and
judge-feedback-driven self-correction loops. These are documented as future work,
not as departures from the architecture.

\end{document}